\documentclass[letterpaper]{article} 
\usepackage{aaai2026}  
\usepackage{times}  
\usepackage{helvet}  
\usepackage{courier}  
\usepackage[hyphens]{url}  
\usepackage{graphicx} 
\usepackage{natbib}  
\usepackage{caption} 
\usepackage{pdflscape}
\usepackage{amsmath}
\usepackage{subcaption}
\usepackage{multicol}
\usepackage{multirow}
\usepackage{pgfplots}
\pgfplotsset{compat=1.18}
\usepackage{xcolor}
\usepackage{booktabs} 
\usepackage{tikz}
\usetikzlibrary{arrows.meta}
\usetikzlibrary{shapes.multipart,positioning, fit} 
\usetikzlibrary{shapes, backgrounds}
\usetikzlibrary{calc} 
\usetikzlibrary{positioning,shapes.multipart,arrows.meta}
\usepackage{threeparttable}
\usepackage{makecell}
\usepackage{multirow}

\usepackage{pgfplots}
\definecolor{myblue}{RGB}{70, 130, 180}     

\tikzset{
  twobox/.style n args={2}{
    rectangle split,
    rectangle split parts=2,
    rectangle split part align=left,
    rounded corners=2mm,
    draw=#1,
    very thick,
    rectangle split part fill={#1!10,white},
    text width=#2,
    inner sep=3pt,
        font=\normalsize,               
    every lower node part/.style={  
      font=\small}
  },
  twobox/.default={black}{6cm},
  arrow/.style={
    -{Stealth[length=2mm]},
    thick
  }
}

\definecolor{myCream}{RGB}{255,253,208} 
\definecolor{myPurple}{RGB}{250,240,250}   
\definecolor{myGreen}{RGB}{230,250,230}    
\colorlet{myPurpleFill}{pink!20!white}
\colorlet{myLighterGreen}{myGreen!60!white} 
\definecolor{myLightBlue}{RGB}{179,229,252}
\colorlet{myBlue}{myLightBlue!60!white}
\definecolor{myPeach}{RGB}{255,229,204} 
\definecolor{OIskyblue}  {HTML}{56B4E9}
\definecolor{Red1}{HTML}{EB3829}
\definecolor{Green1}{HTML}{9FCE63}
\definecolor{Blue1}{HTML}{51AEEA}
\definecolor{orangered}{RGB}{255,69,0} 

\usepackage{algorithm}
\usepackage[]{algpseudocode} 
\algrenewcommand\algorithmicrequire{\textbf{Input:}}
\algrenewcommand\algorithmicensure {\textbf{Output:}}
\algrenewcommand\algorithmicreturn{\textbf{Return}}
\algrenewcommand\algorithmiccomment[1]{\hfill\(\triangleright\)~#1} 

\algrenewcommand\algorithmicindent{1.5em}
\renewcommand{\Call}[2]{\texttt{#1}(#2)}

\title{Contextual Value Alignment via Multilayer Combinatorial Fusion}

\author{
Yuanhong Wu$^{\star}$, Djallel Bouneffouf$^{ \dagger}$, and D. Frank Hsu$^{\star}$
}
\affiliations{
    $^{\star}$  Dept. of Computer and Information Science, Fordham
        University, New York, NY, USA \\
      $^{\dagger}$ T. J.  Watson Research Center, IBM Research, Yorktown  
        Heights, NY, USA
}

\usepackage{bibentry}

\begin{document}

\maketitle

\begin{abstract}

Aligning large language models (LLMs) with human values remains a major challenge, especially for trustworthy AI. While existing approaches such as RLHF, CAI, and their variants have achieved promising results, they often rely on a single-agent framework and a unified reward system. This limits their ability to capture ethical pluralism, adapt to diverse moral contexts, and reflect the dynamics of multi-agent moral reasoning.

In this work, we propose a framework that utilizes multilayer combinatorial fusion for contextual value 
alignment (MCF-CVA). At the first layer of the framework, it instantiates multiple moral agents, each fine-tuned to represent a distinctive value. Their outputs are then expanded combinatorially using both score- and rank-combinations as well as average and weighted aggregations. These combined models are then reduced to the same number of initial moral agents. This expansion and reduction (EAR) process continues for multi-layers until a stopping criterion is reached.

The MCF-CVA framework leverages cognitive diversity between agents to mitigate conflicts and redundancies across multiple agents, producing responses that better reflect contextual human values. The framework using the EAR algorithm is performed on the dual architecture of Euclidean score space and Kemeny rank space. Empirical evaluations demonstrated that the proposed framework outperforms single-agent baselines, multi-agent single-layer results, and previous aggregation approaches on standard metrics, showing that the MCF-CVA framework provides a robust and effective mechanism for advancing contextual value alignment in LLMs.

\end{abstract}

\section{Introduction}

The \emph{value alignment problem}---ensuring that autonomous systems act in accordance with human moral principles---has emerged as one of the most pressing challenges in the development of trustworthy AI~\cite{russell2015research, ji2025ai}. Traditional approaches have treated alignment primarily as a form of \emph{preference inference}, attempting to derive human values from behavioral data or human feedback. However, as artificial agents become more embedded in social and moral contexts, it is increasingly evident that human morality is \emph{pluralistic, contextual, and dynamic}~\cite{gabriel2020artificial}. A single normative model cannot adequately represent the diversity of moral reasoning found across individuals, cultures, and situations.

A wide range of approaches have been developed to align large language models with human values \cite{norhashim2024measuring, singh2025ai, pappas2025human, ji2025ai, li2026landscape}. For example, reinforcement learning from human feedback (RLHF) trains a reward model on human preference data and then optimizes the policy against this learned reward \cite{ouyang2022training}. Constitutional AI (CAI) aligns models by enforcing a fixed set of normative principles, using those principles to guide self-critique and AI-generated feedback during training \cite{bai2022constitutional}. However, these methods share a key limitation: they all treat value alignment as a single-agent, single-objective problem. This design cannot leverage the diversity between distinct moral perspectives, nor can it explicitly represent or reason about disagreements among them. Yet human values are pluralistic and context-dependent: different situations prioritize different values, and different individuals or communities weight these values in different ways. A single agent with a fixed objective is therefore ill-suited to capture this structured pluralism or to support context-sensitive value trade-offs.

To address this limitation, \citet{wu2026enhancing} proposed a single-layer multi-agent value alignment framework, Value Alignment System using Combinatorial Fusion Analysis (VAS-CFA). In this paper, we extend the VAS-CFA to a multi-layer value alignment framework, Multi-layer Combinatorial Fusion for Contextual Value Alignment (MCF-CVA). 
After instantiating a set of moderate size with diverse moral agents, each aligned with a specific moral value, we apply CFA to quantify the diversity between these agents and fuse their generated responses. MCF is then used to perform multi-layer combinatorial fusion using the expansion and reduction (EAR) algorithm 
\cite{zhong2019combining}. The MCF-CVA framework yields responses that are more faithfully aligned with human values than other single layer results.

Combinatorial Fusion Analysis (CFA), proposed by \citet{hsu2006combinatorial, hsu2024combinatorial}, is a robust model fusion paradigm that combines multiple scoring systems by systematically varying combination size and weighting schemes. A key feature of CFA is its notion of cognitive diversity, a diversity measure that leverages both scores and ranks to quantify the dissimilarity in how different scoring systems represent information. Cognitive diversity has been shown to be highly informative for combining diverse models in practice \cite{hsu2019cognitive}. Multilayer Combinatorial Fusion (MCF) extends CFA by stacking multiple layers of CFA combinations to form deeper fusion architectures, thereby more fully exploiting the potential of model fusion across hierarchical layers.

Moral Foundations Theory (MFT) is a widely used tool for modeling moral pluralism and systematic variation in moral judgments across individuals, cultures, and political groups. It was developed by \citet{haidt} to explain how multiple “intuitive ethics” give rise to diverse but patterned moral concerns. In its original formulation, MFT proposed five core moral foundations—Care, Fairness, Loyalty, Authority, and Sanctity. This five-factor structure has been widely applied and shows good psychometric support \cite{graham2011mapping, graham2009liberals}. Later work proposed Liberty as a candidate sixth foundation. However, the Liberty foundation was introduced later as a candidate sixth foundation rather than part of the original core set, and its status as a distinct foundation remains under active discussion \cite{graham2013moral}. Much empirical work—especially factor-analytic and cross-cultural validation studies of the Moral Foundations Questionnaire—continues to adopt and support the original five-foundation structure \cite{de2023moral, dougruyol2019five}. In this paper, we therefore adopt the original five foundations as a well-established and widely used moral basis, and we utilize them by training five independent moral agents, each aligned with one foundation (Care, Fairness, Loyalty, Authority, Sanctity), which serve as the base experts that our MCF-CVA framework later combines.

Multilayer Combinatorial Fusion for Contextual Value Alignment (MCF-CVA) introduced in this paper is a framework that utilizes multi-agent moral reasoning through multi-layer iterative fusion and refinement. At the first layer, MCF-CVA instantiates multiple moral agents, each fine-tuned to represent a distinct moral foundation---such as authority, care, fairness, loyalty, and sanctity. Their outputs are combined using CFA with both score-based and rank-based fusion methods within a dual space of \emph{Euclidean (score)} and \emph{Kemeny (rank)} representations. The system then applies an Expansion-and-Reduction (EAR) algorithm to iteratively fuse, evaluate, and prune agents based on their diversity strength, repeating this process across layers until it stables.

Through this multilayer iterative process, MCF-CVA evolves toward moral decisions that are simultaneously contextually coherent and diversely informed. Our experiments demonstrate that MCF-CVA consistently outperforms single-agent and single-layer baselines across standard metrics, offering a principled approach for reconciling heterogeneous moral reasoning within large language models.

In summary, this work makes the following contributions:

\begin{enumerate}
  \item A novel framework MCF-CVA was proposed to fuse multiple moral agents across multiple iterative reasoning layers by constructing multiple scoring systems corresponding to the multiple moral agents, respectively.
  \item Cognitive diversity was used to measure diversity between the scoring systems (moral agents).
  \item An integration of combinatorial fusion analysis (CFA) and the expansion and reduction (EAR) algorithm to facilitate adaptive multilayer aggregation in both Euclidean score space and Kemeny rank space.
  \item It was demonstrated that the MCF-CVA framework improves moral coherence and is robust and effective over prior contextual alignment methods.
\end{enumerate}

In the following sections, we include: Related Work, Multilayer Combinatorial Fusion, MCF-CVA Workflow, Results, Discussion, and  Limitations followed by Conclusion. Six appendices are also
included: (A) Rank-score function and cognitive diversity, (B) Kemeny rank space as the MCF-CVA architecture, (C) Figure skating judgement, (D) EAR algorithm and the MCF-CVA framework, (E) Performance summary w.r.t. question $q_{2835}$, and  (F) A detailed example of response decomposition in MCF-CVA workflow. 

\section{Related work}

The value alignment problem \citep{russell2015research} aims to design AI systems whose goals and behavior stay aligned with human values. For large language models, one common strategy for value alignment is to fine-tune a single policy with reinforcement learning from human feedback: for example, \citet{wu2023fine} propose Fine-Grained RLHF, which learns reward models from detailed human annotations (toxicity, factuality, relevance, etc.) and then optimizes the model against this scalar reward signal to reduce harmful or low-quality generations. Another widely adopted algorithm is direct preference optimization (DPO) \cite{rafailov2023direct}, where a single language model is trained to prefer human-preferred answers over rejected ones via a contrastive objective, thereby aligning the model to human preferences without an explicit reward model or RL loop.

Beyond single-policy approaches, some work uses multiple human or model sources in value alignment. \citet{noothigattu2018voting} learn a probabilistic model of societal preferences from large-scale Moral Machine data and, at decision time, apply a swap-dominance voting rule to pick the action that best reflects the learned social welfare function, effectively aggregating many human “voters” into one ethical decision. \citet{du2023improving} use a multi-agent debate framework where several language model instances independently generate answers and reasoning, then critique and debate each other over multiple rounds before converging on a final response. And \citet{pang2024self} introduce a self-alignment framework where a single LLM performs multi-agent role-playing simulations to model social interactions and consequences, then uses these simulated critiques to revise its own responses.

\citet{dognin2025contextual} propose contextual moral value alignment, training several agents each associated with different moral values and learning a context-dependent aggregator that selects or combines their responses based on a moral profile for the user. This contextual aggregation idea inspires our work. However, existing multi-source approaches aggregate signals without explicitly measuring or exploiting diversity among moral agents. They do not quantify how differently agents represent moral information. Our MCF-CVA framework, as an extension to the VAS-CFA framework \cite{wu2026enhancing}, builds on this approach by quantifying diversity between moral agents and using multilayer combinatorial fusion to systematically exploit that diversity when fusing their outputs.


 Unlike prior ensemble or alignment techniques, our framework explicitly quantifies and leverages diversity strength across moral agents, applying an Expansion–Reduction process that iteratively fuses and prunes models across both score and rank spaces.

\section{Multilayer Combinatorial Fusion}

Multilayer combinatorial fusion (MCF) employs the expansion and reduction (EAR) algorithm multiple times until it stables. The expansion step uses the CFA  while the reduction step utilizes a sliding rule to select top combined models with large cognitive diversity. This section consists of: (a) Combinatorial fusion analysis, and (b) multilayer combinatorial fusion (MCF) with the EAR algorithm.

\subsection{Combinatorial Fusion Analysis}
 \label{sec:cfa}

 Combinatorial Fusion Analysis (CFA) provides methods and workflows for combining multiple scoring systems (MSS) (e.g., multiple classifier systems, multiple expert systems, multiple language models, and multiple agent systems) in computational learning and modeling, informatics, and intelligent ML/AI systems \cite{hsu2006combinatorial, hsu2024combinatorial}. CFA characterizes a scoring system $A$ with a score function $s_A$, a derived rank function $r_A$, and a function $f_A$ that relates normalized score values to rank values.

Let \(A\) be a scoring system on the dataset \(D=\{d_1,\ldots,d_n\}\). Let \(s_A:D\to R\) be a score function. Rank function \(r_A:D\to N\) is derived by sorting the score values and assigning an increasing rank value to the data item in \(D\) on the decreasing score values. Rank–score function (RSF) \(f_A:N\to R\) is defined as \(f_A(i)=s_A(r_A^{-1}(i))=(s_A \circ r_A^{-1} )(i)\). See Figure \ref{fig:two-tikz}(a) in Appendix A for the relationship between these three functions for a scoring system A \cite{hsu2006combinatorial}. For scoring systems \(A\) and \(B\), cognitive diversity (CD) between $A$ and $B$, \(CD(A,B)\), is defined as the difference between $f_A$ and $f_B$ \cite{hsu2019cognitive}: \(CD(A,B)=d(f_A,f_B)=\left(\sum_{i=1}^{n}(f_A(i)-f_B(i))^2/n\right)^{1/2}\). In addition, the diversity strength of the scoring system \(A\), \(DS(A)\), is the average of \(CD(A,A')\) between $A$ and all other scoring system \(A'\) under consideration. Cognitive diversity CD(A,B) between scoring systems A and B measures the diversity between A and B by calculating the difference between rank-score functions $f_A$ and $f_B$. It is analogous to, but different from, Pearson correlation ($p$), Spearman rho ($\rho$) correlation, and Kendall tau ($\tau$) correlation in statistics. Figure \ref{fig:two-tikz}(b) in Appendix A depicts their relationships and comparisons \cite{hsu2019cognitive, hurley2020multi}. Note that the score function is defined in Euclidean space, whereas the rank function—which handles rankings with or without ties—operates in the Kemeny rank space. Further details on the Kemeny rank space are provided in Appendix B.

The four combination schemes (average score combination (ASC), weighted score combination by diversity strength (WSCDS), average rank combination (ARC), and the weighted rank combination by diversity strength (WRCDS) are defined for each $t \in \{2, 3, 4, 5 \}$ as below, respectively. 
\begin{align}
    s_{ASC}(d_i) &= \frac{\underset{A_j\in \mathcal{A}^\prime}{\sum}s_{A_j}(d_i)}{t} \label{eq:formula 1}\\ 
    s_{WSCDS}(d_i) &= \frac{\underset{A_j\in \mathcal{A}^\prime}{\sum}s_{A_j}(d_i)\, \cdot\, DS(A_j)}{\underset{A_j\in \mathcal{A}^\prime}{\sum} DS(A_j)} \label{eq:formula 2}\\
    s_{ARC}(d_i) &= \frac{\underset{A_j\in \mathcal{A}^\prime}{\sum}r_{A_j}(d_i)}{t} \label{eq:formula 3}\\ 
    s_{WRCDS}(d_i) &= \frac{\underset{A_j\in \mathcal{A}^\prime}{\sum}r_{A_j}(d_i)\, \cdot\, \frac{1}{DS(A_j)}}{\underset{A_j\in \mathcal{A}^\prime}{\sum}\frac{1}{DS(A_j)}}  \label{eq:formula 4}
\end{align}
where $\mathcal{A}^\prime$ is any subset of the full scoring system set $\mathcal{A} = \{A_1, \cdots, A_5\}$ containing at least 2 elements, i.e., $\mathcal{A}^\prime \subseteq \mathcal{A}$, and $\lVert \mathcal{A}^\prime \rVert = t$, with $t \in \{2,3, 4, 5\}$, $s_{A_j}(d_i)$ and $r_{A_j}(d_i)$ are the score and rank assigned by scoring system $A_j$ to the data item $d_i$, respectively; $DS(A_j)$ represents the diversity strength of scoring system $A_j$.

 CFA has been applied to a wide variety of domain applications in scientific discovery and decision making, including more recent results in wireless network selection \cite{kustiawan2017vertical}, material science \cite{tang2021improving}, drug discovery \cite{jiang2023enhancing}, sentiment analysis \cite{patten2025enhancing}, cybersecurity \cite{alfatemi2025identifying, owusu2025generalizable}, Bitcoin price prediction \cite{wu2025bitcoin}, and text classification \cite{xu2025enhancing}.
 It was shown that rank combination tends to perform better than score combination w.r.t. larger cognitive diversity \cite{frank2005comparing}. An example of figure skating judgment is included in Appendix C to illustrate score function, rank function, rank--score function of a scoring system $A$, as well as  score and rank combinations, and cognitive diversity between two scoring systems, $A$ and $B$. 

\subsection{MCF with the EAR algorithm}
\label{sec:mcf}

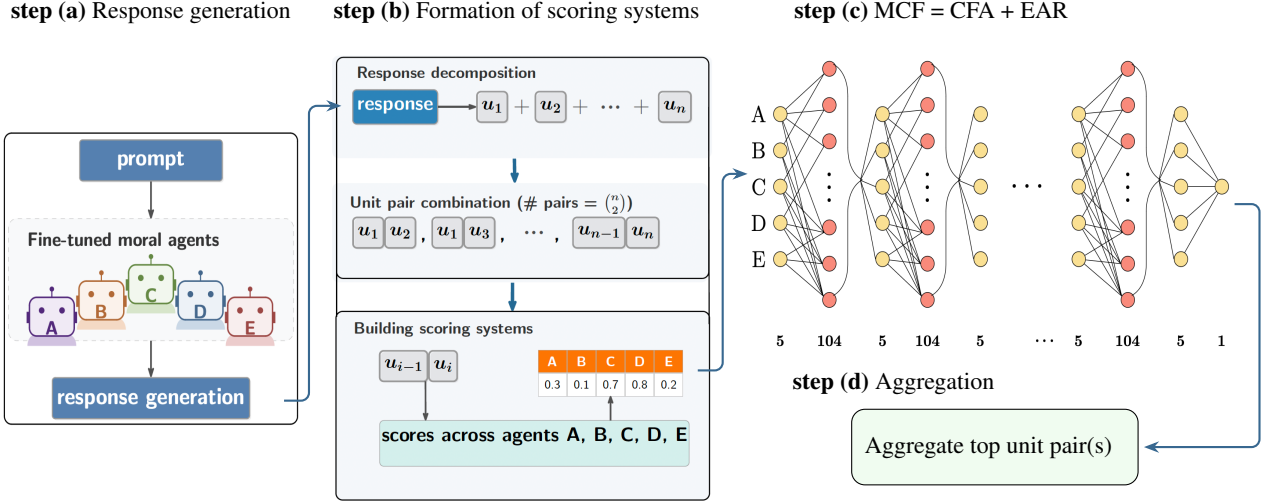
\begin{figure*}[pbt]
    \centering
\begin{tikzpicture}[
        stepbox/.style={
            rectangle,
            rounded corners,
            align=center,
            inner sep=0pt,
        },
        arrow/.style={
            ->,
            >={Stealth[length=2mm]},
            thick,
            shorten <=2pt,
            shorten >=2pt,
            draw=myblue!80!black,
            rounded corners=3pt 
        },
        labeltext/.style={
            align=left,      
            font=\small,    
            anchor=north,      
            yshift=-0.2cm      
        },
        node distance=0.3cm 
    ]

        
        \node[stepbox] (step1) {
            \includegraphics[height=4cm, width=4cm]{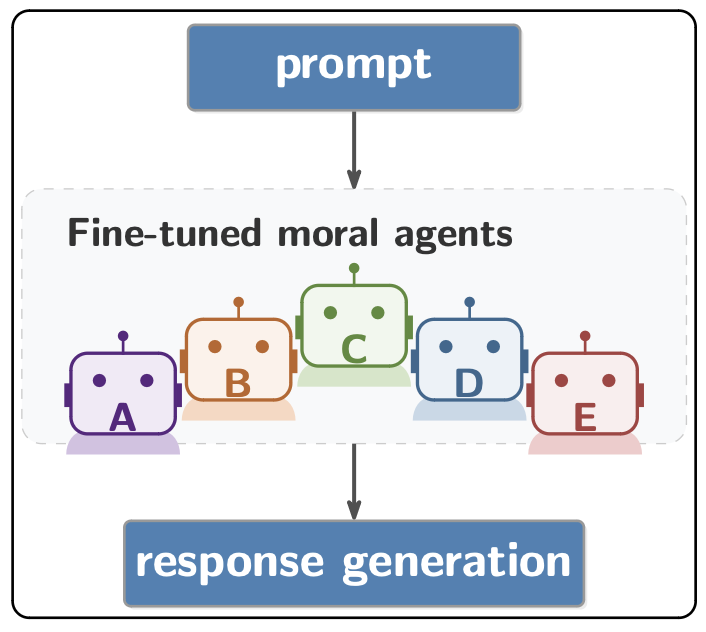}
        };

        \node[stepbox, right=of step1] (step2) {
            \includegraphics[height=6cm]{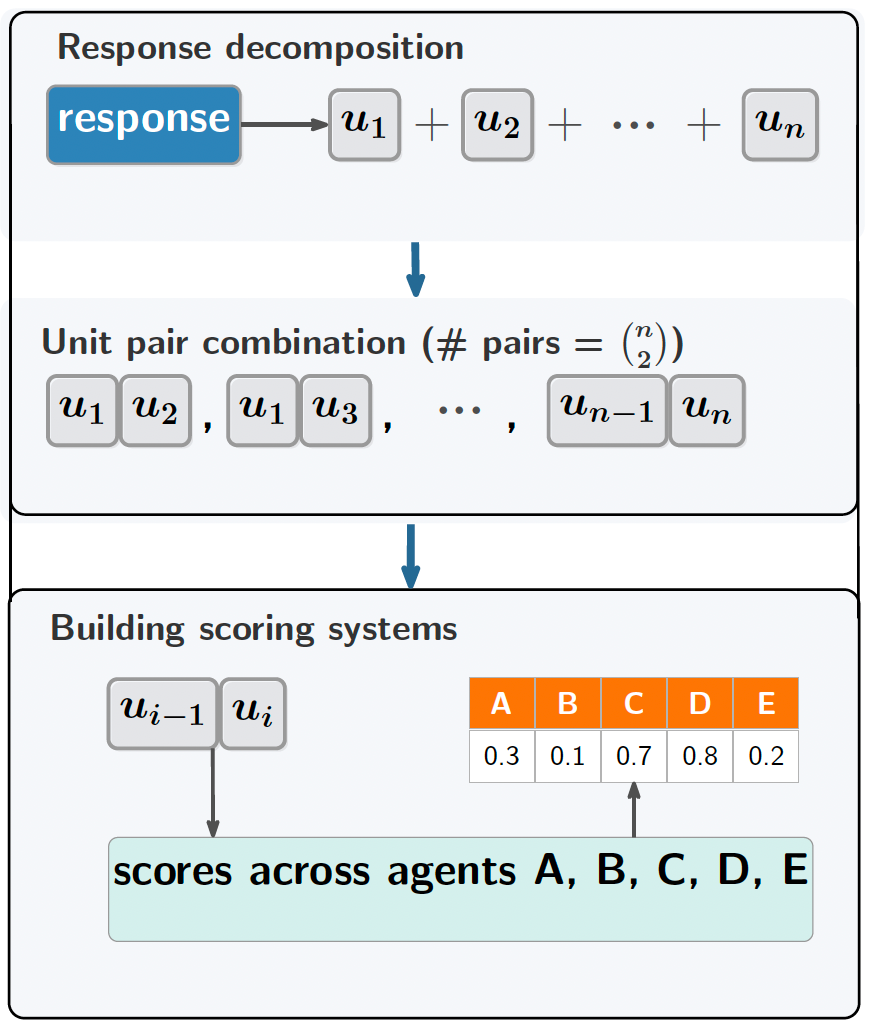}
        };

        \node[stepbox, right=of step2, yshift = 1cm] (step3) {
            \includegraphics[width = 6.5cm, height = 4cm]{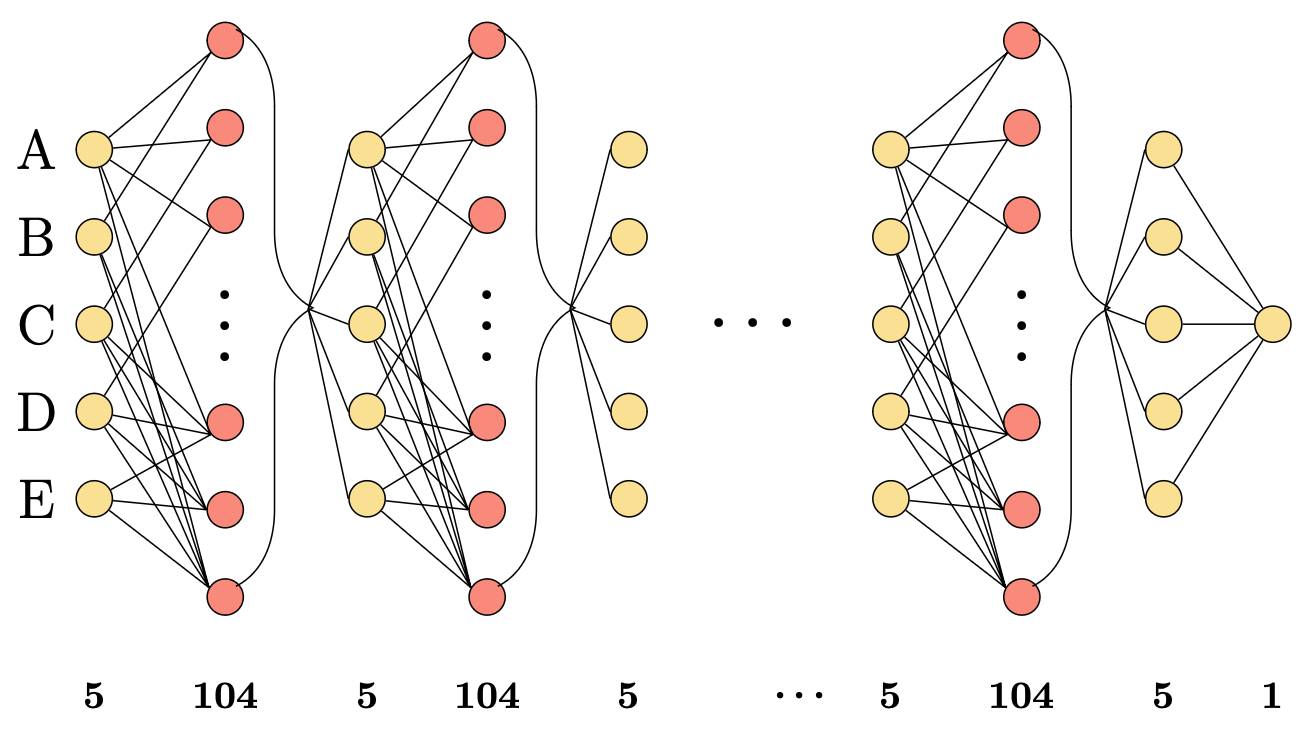}
        };

        \node[stepbox, below=0.7cm of step3, rounded corners=5pt, draw, fill=myLighterGreen, inner sep=5pt, minimum height=1cm] (step4) {
            \makecell{
                {\small Aggregate top unit pair(s) } \\ 
            }
        };

        
        \draw[arrow] ([yshift = -1.6cm, xshift = -0.4cm]step1.east) 
            -- ++(0.4,0) 
            |- ([yshift=2.3cm, xshift = 0.2cm]step2.west); 

        \draw[arrow] ([yshift=-1.2cm, xshift = -0.4cm]step2.east) 
            -- ++(0.4,0) 
            |- ([xshift = 0.1cm, yshift = 0.4cm]step3.west);

       \draw[arrow] ([yshift = 0cm, xshift = -0.2cm]step3.east) 
        -- ++(0.4,0) 
        |- ([yshift=0cm, xshift = 0cm]step4.east);

        \coordinate (labelLine) at (0, 4);

        \node[labeltext] at (step1.center |- labelLine) {
            \textbf{step (a)} Response generation
        };

        \node[labeltext] at ($(step2.center |- labelLine)+(-0.1cm, 0)$) {
            \textbf{step (b)} Formation of scoring systems
        };

        \node[labeltext] at ($(step3.center |- labelLine)+(-0.8cm, 0)$) {
            \textbf{step (c)} MCF = CFA + EAR
        };

        \node[labeltext] at ($(step4.north) + (-1.3cm, 0.8cm)$) {
            \textbf{step (d)} Aggregation
        };

    \end{tikzpicture}
    \caption{The workflow for the MCF-CVA framework.}
    \label{fig:diagram}
\end{figure*}

Multilayer combinatorial fusion employs the expansion and reduction (EAR) algorithm at each layer until it reaches to the layer where there is almost no diversity between models. The pseudocode for EAR algorithm is in Algorithm \ref{alg:er} (Appendix D).We first construct the set of models to be combined. With five models, the total number of nontrivial model subsets is $2^{5} - 5 - 1 = 26$. Since each scoring system consists of both a score function and a rank function, we then extract the corresponding score and rank functions from each system separately. For each input type (scores or ranks), we aggregate them using either average combination or weighted combination by diversity strength, as defined in formulas (\ref{eq:formula 1})-(\ref{eq:formula 4}).

\begin{table}[tpbh]
    \centering
    \caption{Number of questions on each layer for MCF.}
    \begin{tabular}{p{3cm}|p{4cm}}
    \hline
    \#Layer & Number of questions \\
    \hline
     Layer 1 & 1552\\
    Layer 2 & 5457 \\
     Layer 3 & 2988 \\
    Layer 4 & 1056 \\
    Layer 5 & 248 \\
    Layer 6 & 74 \\
    \hline
    Total & 11375 \\
     \hline
    \end{tabular}
    \label{tab:unsupervised_num_layer}
\end{table}

\begin{table*}[hpbt]
    \centering
    \caption{Performance summary for the best single models and the best combined models for layer 1,2,3,4 and 5 on question $q_{822}$, respectively. (\texttt{a}: ASC; \texttt{b}: WSCDS; \texttt{c}: ASC\&WSCDS; \texttt{d}: ARC, \texttt{e}: WRCDS; \texttt{f}: ARC\&WRCDS; \texttt{g}: all four types of combination.)}
    \begin{tabular}{c|l|p{7cm}|c}
    \hline
       \#Layer  &  Model type &   Best model  & F1 BERTScore \\
       \hline
      \multirow{2}{*}{Layer 1} & single & B & \textbf{0.8582}\\
      & combined & CE (\texttt{c}), ACE (\texttt{c}), CDE (\texttt{c}), ACDE (\texttt{c}) & 0.8621	\\ 
      \hline
      \multirow{2}{*}{Layer 2}  & single & D, E & 0.8594\\
      & combined & AD (\texttt{g}), AE (\texttt{g}), DE (\texttt{g}), ACD (\texttt{e}), ACE (\texttt{f}), ADE(\texttt{g}), ACDE (\texttt{f})  & \textbf{0.8659}\\ 
      \hline
    \multirow{2}{*}{Layer 3} & single  & A, C, D, E & 0.8521 \\ 
     & combined & AB (\texttt{f}), DE (\texttt{g)} & 0.8596\\ 
      \hline
    \multirow{2}{*}{Layer 4} & single  & B, C, D, E & 0.8521 \\ 
     & combined & AD (\texttt{c, e}), AE (\texttt{c}), BC (\texttt{g}), ADE (\texttt{b}) & 0.8596\\ 
      \hline
        \multirow{2}{*}{Layer 5} & single  & A, B, C, D & 0.8596 \\ 
     & combined & CD (\texttt{g}), BCD (\texttt{d}) & \textbf{0.8659}\\ 
      \hline
    \end{tabular}
    
    \label{tab:perf_q822}
\end{table*}

The expansion step employs CFA to score combination and rank combination as well as average combination and weighted combination of the initial (or previous) layers to obtain $f(t)=(2^t-1-t)\times 4$ models. The reduction step uses the sliding rule to reduce the $f(t)$ models to $t$ models based on diversity strength at that layer. This process continues to the next layer until there is almost no diversity between models.

\section{MCF-CVA workflow}

We propose a new framework that integrates Multilayer Combinatorial Fusion into context value alignment (MCF-CVA). Figure \ref{fig:diagram} shows the main steps for the proposed framework. 

In \textbf{step (a)} of our MCF-CVA workflow, we fine-tuned five value-specific agents--Authority (A), Care (B), Fairness (C), Loyalty (D), and Sanctity (E)--starting from the OpenAssistant/oasst-sft-4-pythia-12b-epoch-3.5 SFT checkpoint, using Direct Preference Optimization (DPO) with QLoRA \cite{dettmers2023qlora} on a single NVIDIA A100-40GB. The Moral Integrity Corpus (MIC) \cite{ziems2022moral} is used for fine-tuning the individual agents. MIC dataset provides a large set of prompt-response pairs of 113.8K with human revised answers and rich ethical annotations. It uses fixed data splits of 91.0K/11.4K/11.4K samples for train/validation/test. Each moral agent was trained independently from the same base model using different subset of the dataset for 1 epoch with $\beta=0.1$, learning rate $1\times10^{-5}$, per-device batch size $2$ and $8$ gradient-accumulation steps. Specifically, each prompt-response pair is annotated with one or more moral value labels. We partition the dataset into five different subsets, each of which corresponds to a moral value. We then fine-tune the same base model independently on each subset, resulting in five moral agents, each aligned with a specific moral value. QLoRA loads the 12B base in 4-bit NF4 and trains only LoRA adapters, enabling practical single-GPU fine-tuning. Each fine-tuned agent in step (a) generates a response for each prompt in the test set.

\textbf{Step (b)} decomposes each agent’s output into \emph{moral units}, where each unit expresses a single moral claim in the form of a phrase or short sentence. The motivation for this decomposition is that a response often contains multiple moral claims, reflecting moral pluralism. By breaking a complex moral response into distinct, stand-alone claims, we can better identify and manage moral conflict while also reducing redundancy. An additional advantage is that the final response length can be controlled by combining different numbers of moral units. This decomposition is performed using \texttt{GPT-4.1-nano}. We then pool all units together, and form all possible unit pairs by concatenating two units. A detailed example of response decomposition into units and the formation of  moral unit pool is provided in Appendix F. The unit pairs will be considered as data items in multilayer combination fusion.  To score each unit pair for its alignment within each of the five moral values, we train a ``moral classifier.'' Concretely, we encode the human-revised answer using SentenceTransformer (all-MiniLM-L6-v2) and train a logistic regression model for multi-label prediction. Given a pair of moral units, the classifier returns five scores, one per moral value, yielding the scoring table with one row at the end of step (b). We stack the tables together for all data items (unit pairs) to form a big scoring table. Interpreting each column in that table as a value-specific scoring system, we obtain five scoring systems w.r.t. the five models A, B, C, D, and E for the five moral agents, respectively. It's important to note that each prompt in the test set yields a set of five scoring systems and the number of data items may vary across prompts.

In \textbf{step (c)}, we fuse the five scoring systems using MCF. The EAR (Expansion-and-Reduction) algorithm first expands the five systems into 104 scoring systems via four combination schemes (ASC, ARC, WSCDS, WRCDS), and then applies a sliding rule to reduce back to five systems (see Algorithm~\ref{alg:er} in Appendix D). Because MCF combines models combinatorially, we enumerate model sets of sizes $k\in\{2,3,4,5\}$. For each $k$, we construct all $\binom{5}{k}$ subsets from the five systems, yielding $10+10+5+1=26$ distinct model sets in total. For each set, we consider both score-based and rank-based combinations. In Algorithm \ref{alg:er}, we extract the score and rank functions, respectively, from each scoring system (each system comprises a score function, a rank function, and a rank--score function). We then apply the four combination methods within each model set, producing four fused models per set. Hence, the expansion stage yields $4\times 26=104$ combined models prior to reduction.

\begin{figure*}[hpbt]
    \centering
    \includegraphics[width=0.3\linewidth]{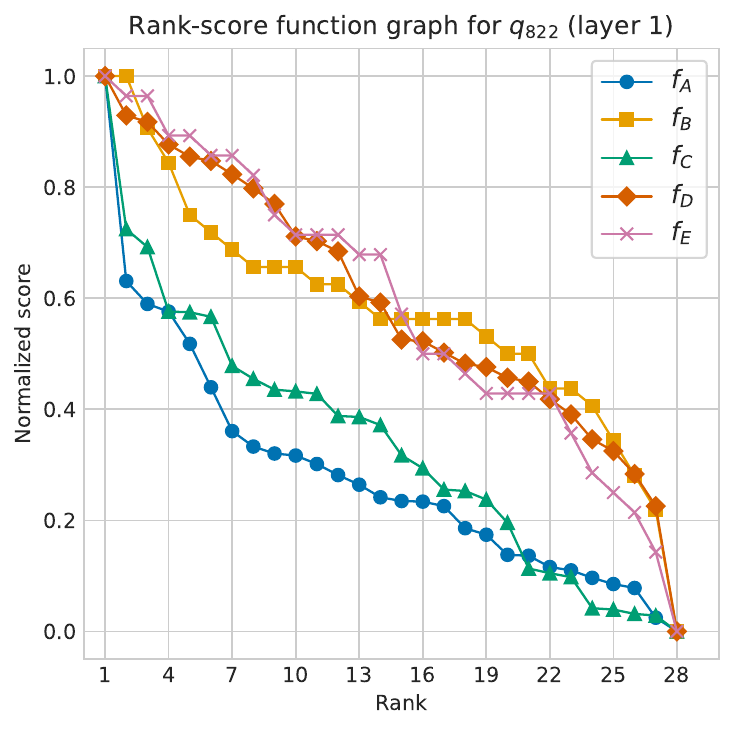}
    \includegraphics[width=0.3\linewidth]{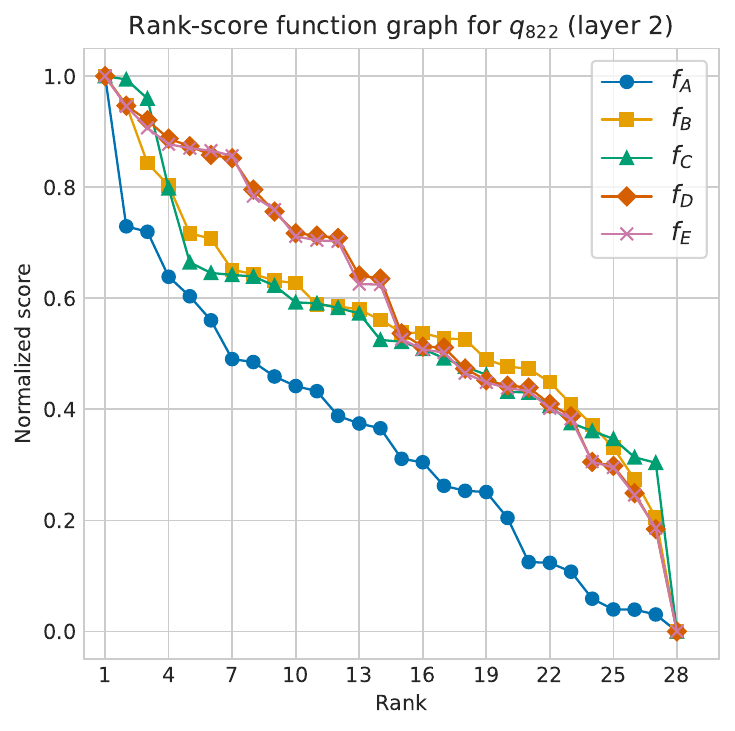}
    \includegraphics[width=0.3\linewidth]{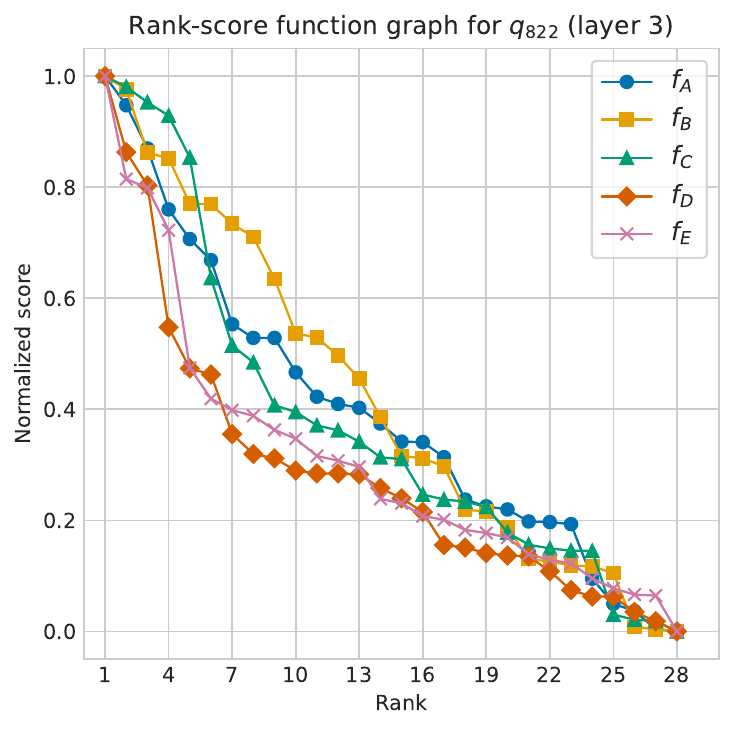}
    \includegraphics[width=0.3\linewidth]{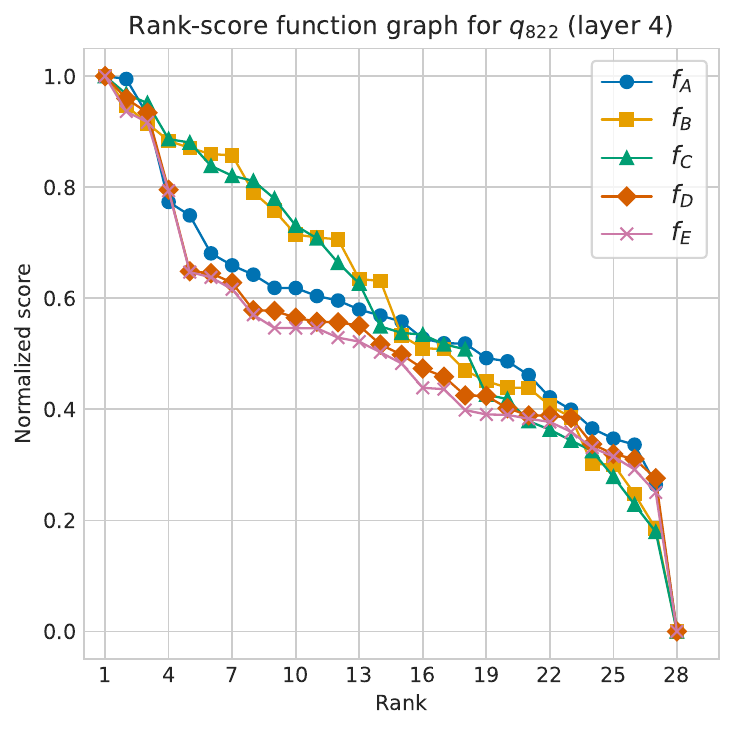}
    \includegraphics[width=0.3\linewidth]{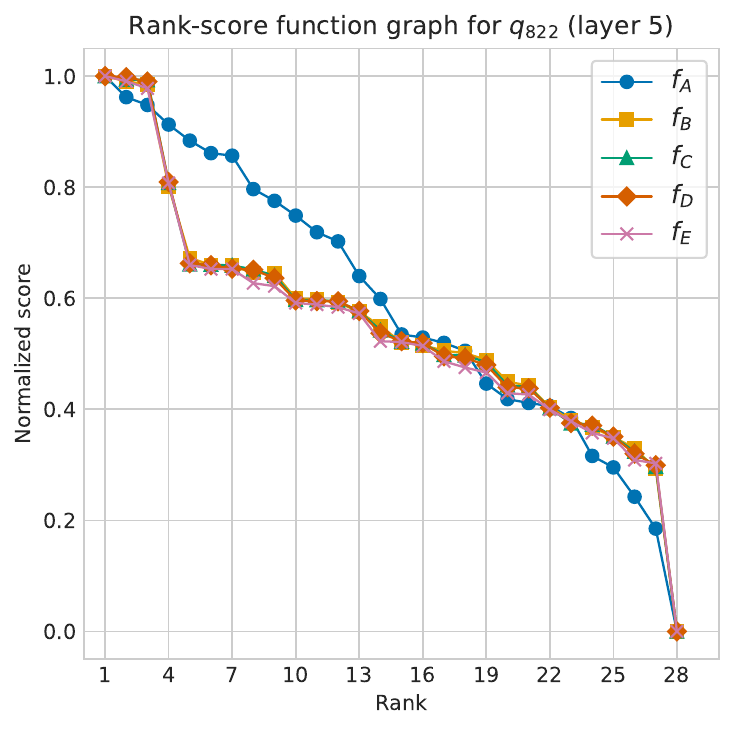}
    \caption{Rank-score function graph for 5 layers on question $q_{822}$.}
    \label{fig:q822}
\end{figure*}

For reduction, we compute each model's diversity strength across the $104$ models. 
To reduce to five models, we rank all $104$ models by diversity strength in descending order. We then apply a sliding-rule selection: starting at the top of ranking, we sweep downward until we obtain five scoring systems that have the highest diversity strength; these constitute the reduced set. We repeat the expansion--reduction cycle on this reduced set until a stopping criterion is met: either the maximum diversity strength falls below 0.05 or the number of layers reaches six. At the final layer, we apply average rank combination to fuse the five models into a single final model. 

In \textbf{step (d)}, this single final model is used for unit aggregation. The number of unit pairs to be aggregated can be adjusted according to the desired response length. For a shorter answer, only the top-ranked unit pair is aggregated, whereas for a longer and more morally rich response, the top two or three unit pairs can be combined. The aggregation is performed by \texttt{GPT-4.1-nano}. Note that sentence decomposition and unit aggregation used the same language model. Since the key part of our framework is the MCF process, we used a small off-the-shelf LLM without additional fine-tuning for both decomposition and aggregation. The resulting response is then evaluated against the ground-truth response using F1 BERTScore. The score obtained is reported as the performance of the MCF-CVA workflow for that question. This process is repeated for every question in the test set. The pseudocode for MCF-CVA is in Algorithm \ref{alg:mcf4va} (Appendix D). 

We note that each data item corresponds to a unit pair, and aggregation over unit pairs yields the final response for a prompt. However, response decomposition typically produces many candidate units, making the choice of which two units to aggregate non-trivial, as units may be redundant or conflicting. MCF-CVA addresses this challenge by providing a systematic mechanism to leverage the diversity of scoring systems—through four distinct combinations—thereby promoting the more informative unit pair.

\section{Results}

In this section, we report results for the MCF-CVA framework: (a) Results evaluated on the Moral Integrity Corpus (MIC) dataset; (b) Ablation results for the framework to show MCF is the key for performance gain; (c) Results of applying MCF to value alignment classification task on the Commonsense dataset.

\subsection{Results on MIC dataset}

Table \ref{tab:unsupervised_num_layer} summarizes the number of questions that stop at each layer. A question is said to stop at a layer if it satisfies the MCF stopping criterion and halts at that layer. Overall, 1,552 questions stop at Layer 1, while 9,823 questions continue beyond the first layer. About half of them (5,457) stop at Layer 2, 2,988 at Layer 3, and 1,056 at Layer 4. And even 248 questions go to Layer 5 while Layer 6 has 74 questions.  These findings indicate that MCF-CVA adapts at the question level: different questions may continue at different depths, which can make MCF more effective than a single-layer CFA. We provide two illustrative examples $q_{822}$ and $q_{2835}$. For each example, we plot the rank–score function graph and the performance of best combined models of each layer. The question $q_{822}$ that stops at Layer 5 appears in Figure \ref{fig:q822}. The second example $q_{2835}$ that stops at Layer 6 is included in Appendix E. From Figure \ref{fig:q822}, we observe how the rank–score functions evolve across five layers; their diversity decreases gradually through the five layers. Table \ref{tab:perf_q822} reports the best performance at each layer for question $q_{822}$, listing both the top individual model(s) and the best combined model(s). The best individual model at the beginning is B with an F1 BERTScore of 0.8582; through multilayer iteration the question stops at Layer 5, achieving 0.8659.

Then, we compare our MCF-CVA model performance against individual moral agents in Table~\ref{tab:compare_all}. The first row reports test performance for the five single agents. The best agent is Loyalty with a score of 0.8663. 

Next, we have some other multi-agent models. On the same dataset, prior multi-agent methods CVA--GS and CVA--GS--DYN report a BERT score of 0.8728 and 0.8754 \cite{dognin2025contextual}, and our previous model single-layer VAS-CVA using a single unit using CFA \cite{wu2026enhancing} has a performance of 0.8849. Multi-agent debate improves value alignment through an iterative debate process. Following \citet{du2023improving}, we use the five single agents as debaters and implement the debate in 2 rounds. We consider this debate as a 2-layer multi-agent model.  In each round, every agent is given the responses of the other four agents as additional context and is asked to revise its own response accordingly. The multi-agent debate framework achieves a performance of 0.8698, which is slightly higher than that of the best individual agent.

We then evaluate the single-layer variant of the MCF-CVA framework. For each question in the test set, we use the model in the first layer (step (c) in Figure \ref{fig:diagram}). The final performance is obtained by averaging over all questions. The single-layer MCF-CVA achieves an accuracy of 0.9007, outperforming all individual base models and prior multi-agent models. We next evaluate our most powerful model, multi-layer MCF-CVA. The F1 BERT score achieved is 0.9098, which is almost $1\%$ performance gain compared to the single-layer counterpart. In sum, MCF-CVA outperforms individual agents, single-layer variant, and prior multi-agent performance.

\begin{figure}[htbp] 
\centering

\begin{tikzpicture}
\begin{axis} [
    width=1.0\columnwidth, 
    height=7.2cm,          
    ybar,
    bar width=0.25,        
    ymin=0.80,
    ymax=0.925,
    ylabel={F1 BERTScore},
    ylabel style={font=\bfseries\large}, 
    xmin=0.625,
    xmax=4.575,              
    xtick={
        1.0, 1.25, 1.50, 1.75, 2.0, 
        2.6, 2.85, 3.10, 
        3.7, 3.95, 4.20
    },
    xticklabels={
        B, A, C, E, D,
        CVA-GS, CVA-GS-DYN, VAS-CFA, 
        {Multiagent Debate}, single-layer variant, MCF-CVA
    },
    xticklabel style={
        font=\scriptsize, 
        rotate=30,        
        anchor=east,
        align=center
    },
    ytick={0.80,0.82,0.84,0.86,0.88,0.90,0.92},
    xtick pos=left,         
    ytick pos=left,         
    tick pos=left,          
    ymajorgrids=true,
    grid style={dashed, gray!35},
    axis line style={black!60},
    tick style={black!60},
    tick align=outside,
    legend cell align=left,
    legend style={
        at={(0.5,1.05)},
        anchor=south,
        legend columns=1, 
        draw=none,
        font=\scriptsize, 
        row sep=0.05cm
    }
]

\draw[dashed, red!80!black, line width=0.8pt] (axis cs:0.5,0.8663) -- (axis cs:5.2,0.8663);
\draw[dashed, red!80!black, line width=0.8pt] (axis cs:0.5,0.8849) -- (axis cs:5.2,0.8849);
\draw[dashed, red!80!black, line width=0.8pt] (axis cs:0.5,0.9098) -- (axis cs:5.2,0.9098);

\addplot[
    draw=blue!70!black,
    fill=blue!55,
    line width=0.5pt,
    bar shift=0pt,
] coordinates {
    (1.0,0.8533)
    (1.25,0.8569)
    (1.50,0.8628)
    (1.75,0.8653)
    (2.0,0.8663)
};

\addplot[
    draw=magenta!60!black,
    fill=magenta!45,
    line width=0.5pt,
    bar shift=0pt,
    forget plot
] coordinates {
    (2.6,0.8728)
    (2.85,0.8754)
    (3.10,0.8849)
};

\addplot[
    draw=green!50!black,
    fill=green!65!black,
    line width=0.5pt,
    bar shift=0pt,
    forget plot
  ] coordinates {
    (3.7,0.8698)  
    (3.95,0.9007)  
    (4.20,0.9098)  
};

\draw[dashed, thick, gray!60] (axis cs:2.30,0.80) -- (axis cs:2.30,0.925);
\draw[dashed, thick, gray!60] (axis cs:3.40,0.80) -- (axis cs:3.40,0.925);

\end{axis}
\end{tikzpicture}

\caption{Comparison of MCF-CVA models with individual models (blue), other single-layer multi-agent fusion models (pink), and the multi-layer multi-agent debate model (green).}
\label{fig:f1_bertscore_bar}
\end{figure}
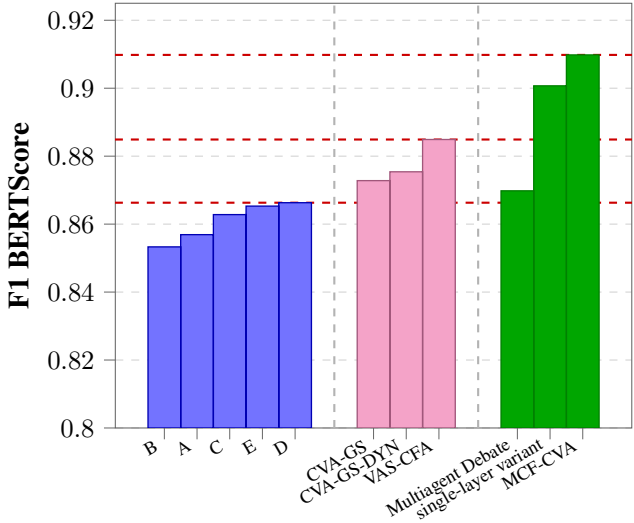

Figure \ref{fig:f1_bertscore_bar} provides a visual comparison of individual models, existing multi-agent fusion baselines, and the proposed MCF-CVA models. The results show a clear progression from individual agents to existing fusion methods and finally to the proposed MCF-CVA models. This visual pattern makes the improvement trend more direct than the table, showing that fusion methods generally improve over single agents, while MCF-CVA further advances the performance beyond existing fusion baselines.

One notable observation from Figure \ref{fig:f1_bertscore_bar} is both Multiagent Debate and the proposed MCF-CVA model are multi-agent, multi-layer frameworks in the third group, but their performance differs substantially. We believe this large difference may be attributed to the underlying architecture.

\begin{table}[htbp]
    \centering
    \begin{threeparttable}
        \caption{Performance comparison of individual agents, prior models and the proposed MCF-CVA models. Agent A, B, C, D, and E refer to authority, care, fairness, loyalty, and sanctity, respectively.}
    \begin{tabular}{p{2.7cm}|>{\centering\arraybackslash}p{2.3cm}|>{\centering\arraybackslash}p{1.8cm}}
    \toprule
   Method &  Model           & F1 BERTScore \\
   \hline 
   \multirow{5}{*}{ Single model}  & A  &  0.8569\\
    &  B       &  0.8533\\
    &  C    &  0.8628\\
    &  D      &  \textbf{0.8663}\\
    &  E    &  0.8653\\
    \hline
    \multirow{2}{*}{\citet{dognin2025contextual}} & CVA-GS & 0.8728\\
                            & CVA-GS-DYN & 0.8754\\
                \hline
    \citet{wu2026enhancing} & VAS-CFA & \textbf{0.8849} \\
\hline
      Multi-agent debate & -- & 0.8698 \\
      \hline
    \multirow{2}{*}{\makecell{MCF-CVA}}  & single-layer &  \textbf{0.9007}\\ 
      &  multi-layer & \textbf{0.9098} \\ 
        \bottomrule
    \end{tabular}
    \label{tab:compare_all}
    \end{threeparttable}
\end{table}

\begin{figure}[hbtp]
    \centering
    \includegraphics[width=\linewidth]{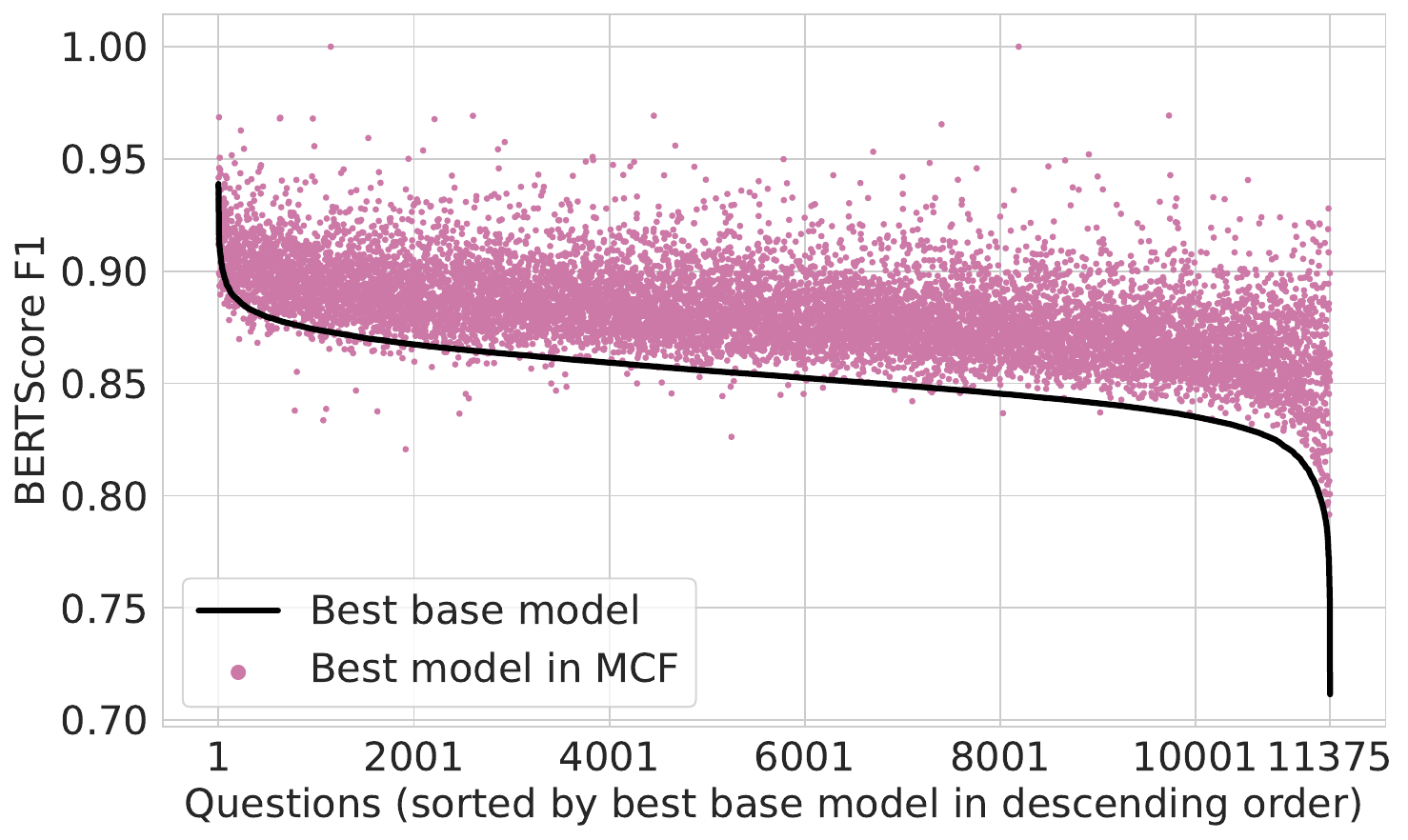}
    \caption{Performance comparison for best base models and best MCF models across all prompts in the test set.}
    \label{fig:base_mcf}
\end{figure}

Finally, we compare the performance of the MCF models with the base models for each question, as shown in Figure \ref{fig:base_mcf}. Each pink point represents the performance of the final MCF model for a single question, and the dataset contains 11,375 questions in total. The black line represents the performance of the best base model on the same question. The questions are sorted in descending order according to the best base-model performance. As shown in the figure, most pink points lie above the black line, indicating that the MCF models outperform the base models for the majority of questions, although a small number of cases fall below the line. This result demonstrates that MCF not only improves the average performance compared with the base models (as shown in Table \ref{tab:compare_all}) but also achieves better performance on most individual questions.

\subsection{Ablation study}

Next, we conduct an ablation study to evaluate the contribution of each component in our framework. We also examine the effect of using different numbers of moral unit pairs. The results are reported in Table 4.

In the original MCF-CVA framework, we use the top 1 unit pair, which contains two moral units, for aggregation and evaluation. However, when the input context is more complex, two moral units may not be sufficient to capture the full moral meaning of the response. Therefore, we also test the setting of using the top 2 unit pairs. As shown in Table 4, using 2 unit pairs decreases the BERTScore from 0.9098 to 0.8936, a drop of about $1.6\%$. However, this result is still relatively strong compared with the previous baselines. One possible reason for this decrease is that the target responses in the MIC test dataset are usually short, suggesting that the moral context is relatively simple and may not require many moral units. Adding more unit pairs may introduce extra information or noise, which slightly reduces performance.

\begin{table}[ht]
\centering
\caption{Ablation study of the proposed framework by removing decomposition, MCF, or aggregation components.}
\label{tab:mcf_units}
\begin{tabular}{p{2.4cm}p{2.6cm}c}
\toprule
\textbf{Model} & \textbf{Moral Units} & \textbf{BERTScore} \\ \midrule

\multirow{2}{*}{MCF-CVA} 
& top 1 unit pair & 0.9098 \\ \cmidrule{2-3}
& top 2 unit pairs & 0.8936 \\ \hline

\multirow{2}{*}{w/o MCF} 
& random 1 unit pair & 0.8827 \\ \cmidrule{2-3}
& random 2 unit pair & 0.8701 \\ \hline

\multirow{2}{*}{w/o aggre.} 
& top 1 unit pair & 0.9075 \\ \cmidrule{2-3}
& top 2 unit pair & 0.8916 \\ \hline

w/o decomp. + MCF 
& -- & 0.8621 \\ 
\bottomrule

\end{tabular}
\end{table}

We then remove the MCF component from the framework. Without MCF, the framework cannot rank and select the most effective moral unit pairs. Therefore, we randomly select 1 or 2 unit pairs for aggregation and evaluation. When randomly selecting 1 unit pair, the BERTScore drops to 0.8827, which is a clear decrease compared with the full MCF-CVA result. When randomly selecting 2 unit pairs, the score further decreases to 0.8701. These results show that MCF plays an important role in identifying high-quality moral unit pairs and improving the final response quality.

Next, we remove the aggregation component. In this setting, the selected unit pairs are evaluated directly without being aggregated into a final response. The performance only slightly decreases from 0.9098 to 0.9075 when using the top 1 unit pair, and from 0.8936 to 0.8916 when using the top 2 unit pairs. This suggests that aggregation contributes to the final framework, but its effect is relatively small compared with the MCF component.

Finally, we remove both decomposition and MCF. Since MCF operates on decomposed moral units, removing decomposition also prevents the use of MCF. In this setting, we directly aggregate the original responses and evaluate the result. The BERTScore drops to 0.8621, which is lower than the best base models. This result indicates that directly combining full responses without moral-unit decomposition and MCF is not effective.

Overall, the ablation study shows that MCF is the key component driving the performance improvement of the proposed framework. Decomposition provides the necessary structure for MCF to operate on moral units, while aggregation further refines the selected units into a smooth final response.

\subsection{Results on Commonsense dataset}

Another common contextual value alignment task, beyond question answering, is value alignment classification: given a conversational context, the goal is to determine whether the response is aligned with human values. Our MCF-CVA framework can be adapted to this type of task. The shared core remains the use of MCF to fuse multiple scoring systems. In this subsection, we report the results on the Commonsense dataset.

The Commonsense Morality subset of the ETHICS benchmark contains scenario descriptions based on widely shared human moral intuitions regarding whether an action is clearly wrong \cite{hendrycks2020aligning}. The dataset includes both short (1–2 sentence) examples collected via Amazon Mechanical Turk and longer, multi-paragraph narratives curated from online forums. Labels were obtained through majority voting among multiple English-speaking annotators. The dataset is provided with a standard train/development split for supervised learning and a held-out test set for evaluation, making it suitable for training and assessing models’ ability to capture commonsense moral norms in open-world contextual scenarios.

The Commonsense dataset has two labels: 0 for not morally wrong (aligned) and 1 for clearly morally wrong ( misaligned), so it is a binary classification problem. The goal is to predict whether an action “clearly should not have been done". We apply MCF to combine five diverse base models: A) Linear SVM, B) XGBoost, C) Random Forest, D) RoBERTa-large, and E) DeBERTa-large. For the traditional ML models (A–C), we encode text using TF–IDF features, split the developement set into train/validation ($10\%$ validation). For the language models (D–E), we fine-tune pretrained transformers by tokenizing text with the model’s own tokenizer. The model diversity is strengthened by the different feature encodings (TF–IDF vs transformer tokenization). Finally, we run the MCF method iteratively until the stopping criterion is met.

\begin{figure*}[htbp]
    \centering
    \includegraphics[width=0.32\linewidth]{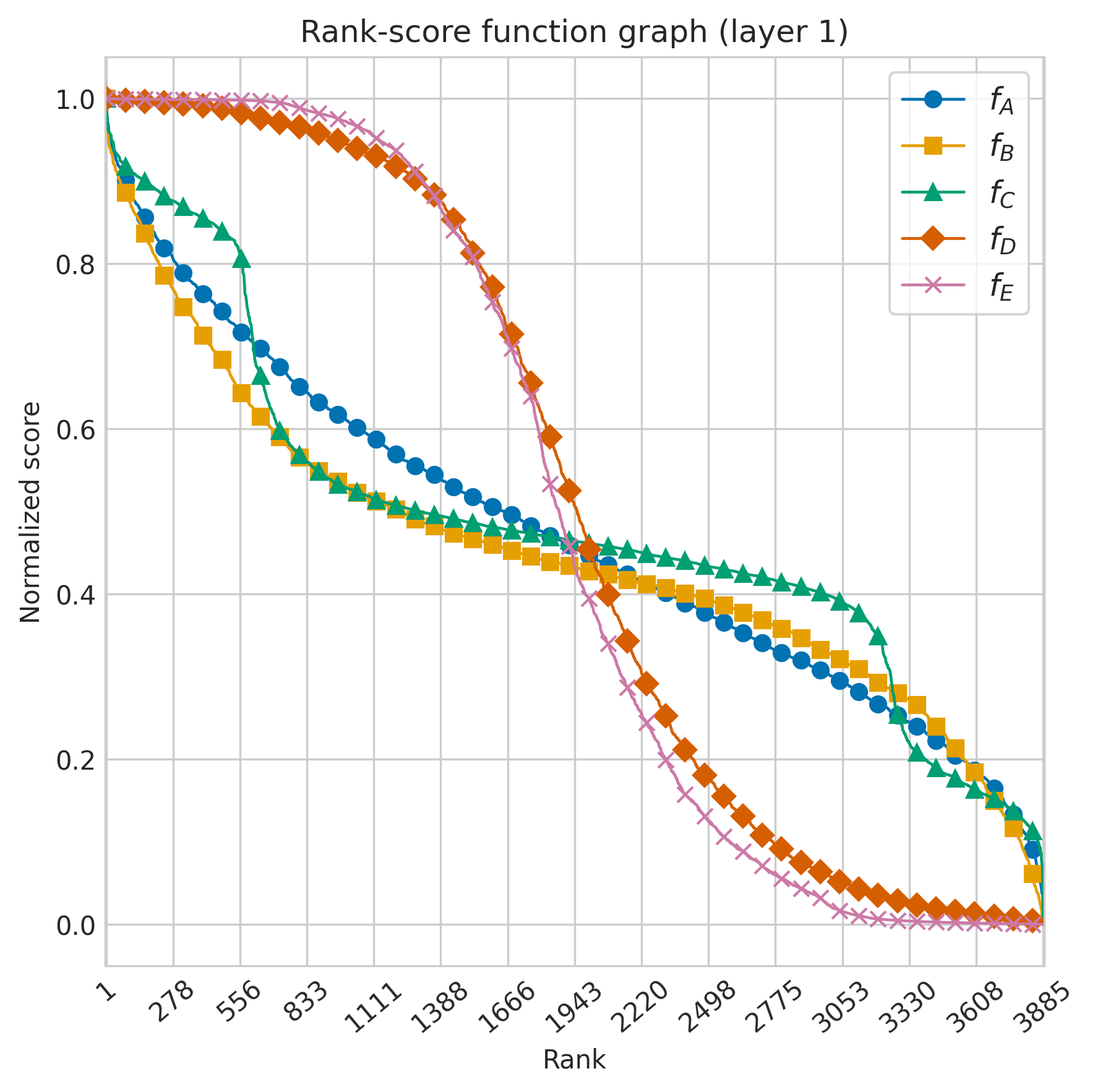}
    \includegraphics[width=0.32\linewidth]{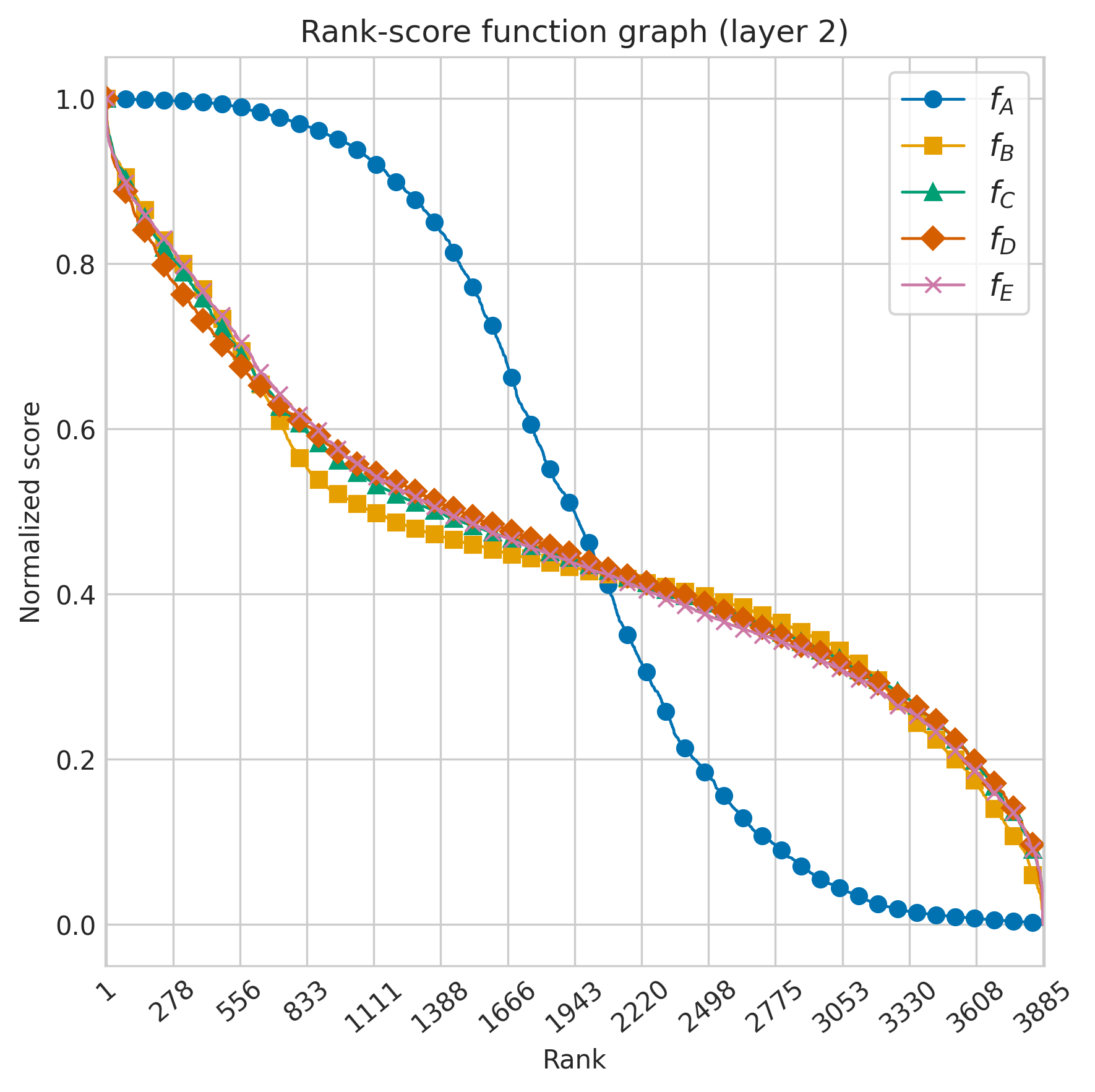}
    \includegraphics[width=0.32\linewidth]{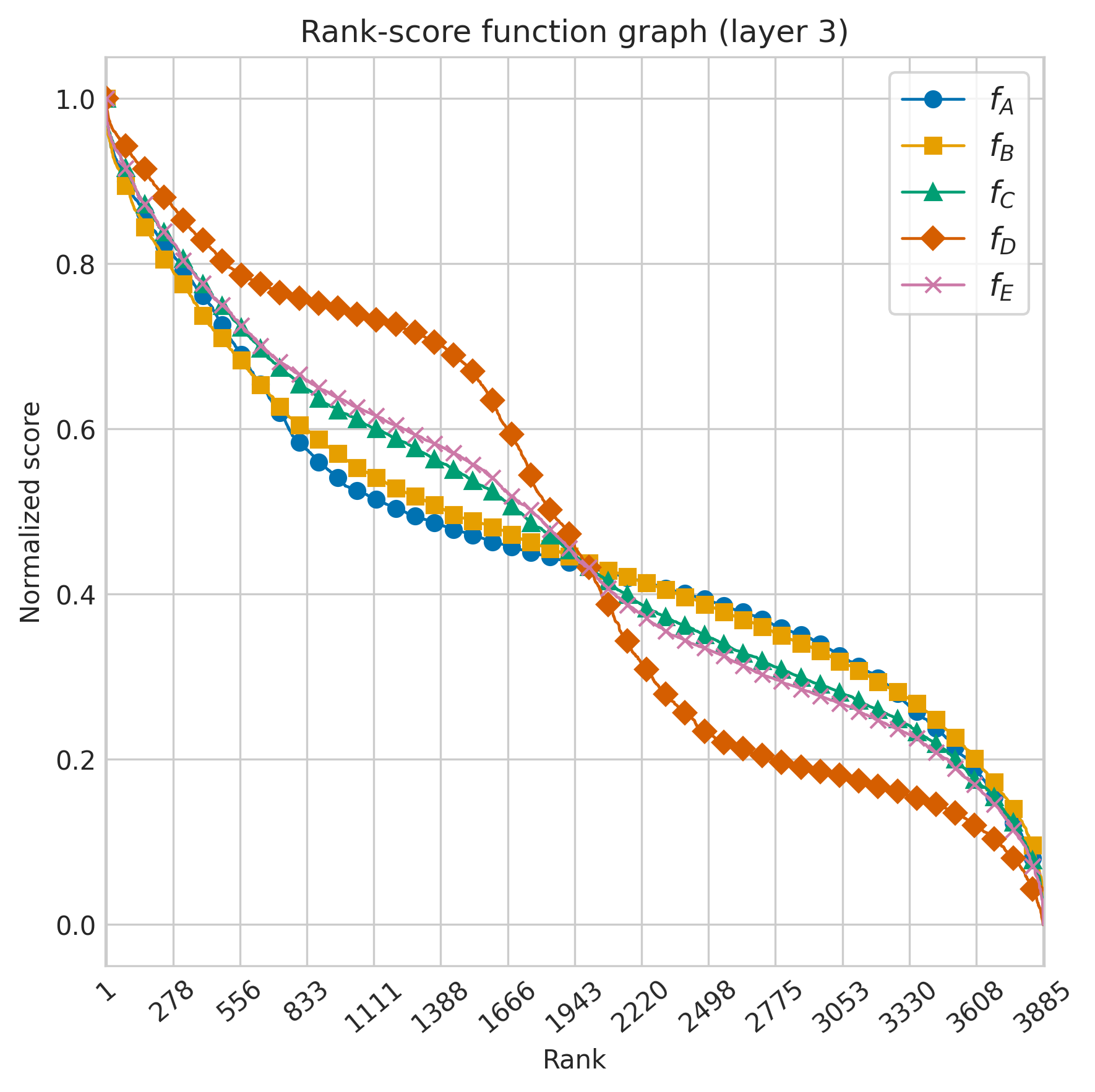}
    \caption{Rank-score function graph for Commensense dataset.}
    \label{fig:rsf_cm}
\end{figure*}

The rank-score function graph for Commensense dataset is shown in Figure \ref{fig:rsf_cm}. We can see the cognitive diversity between models in each layer shrinks across layers. The layer-1 rank-score function graph of model D (RoBERTa-large) and E (DeBERTa-large) are very close to each other while models A, B, and C cluster together. This indicates strong diversity between traditional ML models and language models, but the diversity within each group is relatively small.

\begin{table}[H]
    \centering
    \caption{Best model performance comparison for each layer on Commensense dataset.}
    \begin{tabular}{p{2.1cm}|p{1.6cm}|p{1.7cm}|p{1.2cm}}
    \Xhline{1pt}
                                 &      Model type                                       & Model                & Accuracy  \\
                                 \hline
       \multirow{2}{*}{Layer 1}  &  single     & E & 0.8970  \\
                                 &  combined & BCE({\scriptsize ASC})  & 0.9331 \\
                                 \hline
        \multirow{2}{*}{Layer 2}  &  single    & A & 0.8988  \\
                                 &  combined &  ABE({\scriptsize ASC})  & 0.9243\\
                                 \hline
        \multirow{2}{*}{Layer 3}  &  single     &   D & 0.9166 \\
                                 &  combined &  CDE({\scriptsize WSCDS}) & \textbf{0.9220} \\ \hline
            \multirow{2}{*}{
\makecell{
\citeauthor{hendrycks2020aligning}\\
(\citeyear{hendrycks2020aligning})
}}
            & \multirow{2}{*}{single} & RoBERTa-large & 0.9040\\
                                                                   &         & BERT-large & 0.8850 \\
    \Xhline{1pt}
    \end{tabular}
    
    \label{tab:perf_cm}

\end{table}

Table \ref{tab:perf_cm} presents the performance for the best base model and best combined model for each layer. We can see the best base model in the first layer is model E with accuracy of 0.8970. The best combined model in the first layer is BCE by average score combination, which achieves accuracy of 0.9331. It's a big improvement. In the second layer, the performance for the best base model is 0.8988 while the performance is 0.9243 for the best combined model. The MCF stops at the third layer with the final accuracy of 0.9220. We compare our final performance on layer 3 with the result from \citet{hendrycks2020aligning}. The final accuracy of 0.9220 exceeds the result reported in prior study, demonstrating the effectiveness of the MCF approach.

Combining all results together, multilayer combinatorial fusion can be effectively applied to both contextual 
value alignment tasks: question answering  and alignment classification. In question answering, responses are split into units to construct scoring systems whereas in classification, the base models can directly output scores. The EAR algorithm is followed to iteratively expand and reduce the scoring systems until the stopping condition is reached.

\section{Discussion}

One important design choice in MCF-CVA is that the framework combines \emph{moral units} rather than full agent responses. A complete response often contains several distinct moral claims. Directly aggregating whole responses may therefore mix together incompatible or repetitive meanings, making it harder to identify which parts of different agents are genuinely complementary. By decomposing a response into smaller stand-alone moral units, MCF-CVA creates a more structured representation of moral content, so that fusion operates on semantically cleaner building blocks. This design is also consistent with the role of decomposition in our workflow: it helps isolate conflict, reduce redundancy, and gives the framework direct control over the final response length through the number of selected unit pairs. In this sense, moral units provide an intermediate granularity that is more informative than tokens but more manageable than complete responses.

A related question is \textbf{why we use unit pairs rather than single units or larger groups such as triples}. The reason is both methodological and practical. In CFA, the scoring systems require a set of data items to compare and rank. If a single unit were treated as one data item, the pool would often be too limited to support robust fusion. At the other extreme, forming all possible triple-unit groups would substantially increase computational cost. More importantly, triple grouping would make response-length control less flexible: selecting one triple already yields three units, while selecting two triples yields six units, which may be too coarse for generating responses of different lengths. Unit pairs provide a useful middle ground. They expand the candidate space enough for meaningful scoring and ranking, but remain computationally feasible. They also offer finer control over final answer length, since longer responses can be produced by aggregating more pairs. Thus, the use of unit pairs is not arbitrary; it reflects a balance between representational adequacy, computational tractability, and controllable generation.

Another central design choice is \textbf{the set of moral values used to instantiate the base agents}. In this work, we adopt the five canonical moral foundations---Care, Fairness, Loyalty, Authority, and Sanctity---because they provide a well-established and widely used basis for modeling moral pluralism, and we already notes that the original five-foundation structure remains the most stable starting point compared with later additions such as Liberty. At the same time, our framework does not assume that more moral values are always better. The goal of MCF-CVA is not simply to maximize the number of base agents, but to select agents with sufficient diversity. Additional values such as Liberty or Honesty could be incorporated in future extensions, but only if they contribute meaningfully different moral perspectives. If newly added values overlap too strongly with existing ones, their orthogonality is limited, and the benefit of adding them may be small. This suggests an important future direction: instead of asking how many moral values can be included, we should ask which subsets of moral values are most diverse, most complementary, and most useful for contextual alignment. More broadly, this raises a deeper question for value alignment research: whether effective pluralistic alignment requires a large inventory of moral values, or whether a smaller but carefully chosen set of sufficiently distinct foundations is already enough to capture most of the practical gains.

One question of great interest in computational learning and ensemble methods is: \textbf{How many base models (or agents, scoring systems) one should use?} Although it may depend on the domain application, an ideal number of base models is $4\leq n\leq 6$ as long as the base models are relatively good (close to the target) and diverse. \citet{wang2001does} demonstrated that the distance between $n$ combined models (scoring systems) and the target sharply decreases at $n$ between 3 and 4. Adding more base models may still improve the performance. But the principle of diminishing return kicks in.

One more issue is \textbf{tie ranking (or tie scoring)}. If a scoring system $A$ on $n$ data items has ties w.r.t.  score function values $s_A$, then the rank function $r_A$ is not a permutation of the $n$ elements $\{1, 2, 3, \cdots\}$. In this case, a ranking with ties is a node in Kemeny rank space $K_n$ (Figure \ref{fig:appA-K3}, Appendix B2), which is a superset of the Bubble sort Cayley graph space $B_n$ (Appendix B1). In such a case, a rank function with ties is represented as a matrix instead of an array \cite{emond2002new}. We note that tie rankings do occur even it's not often. In current work, we resolved tie ranking by taking average of the rank numbers. Hence, CFA and MCF are performed in the Bubble sort Cayley graph space (Appendix B1). 

\section{Limitations}

While the results of MCF-CVA are encouraging, several limitations remain and point to important directions for future work. First, the core iterative mechanism of our framework, namely the Expansion-and-Reduction (EAR) algorithm, is currently supported primarily by empirical evidence rather than a full theoretical analysis. In the present work, the multilayer process stops when either the maximum diversity strength falls below 0.05 or the number of layers reaches six. Empirically, we observe that diversity usually decreases as the iterations proceed, and this trend can be visualized through the rank-score function graphs across layers. However, we do not yet have a closed-form characterization of the EAR dynamics, nor a formal analysis of how diversity strength evolves over iterations or under what conditions convergence behavior can be guaranteed. 

A second limitation concerns evaluation. As an initial study of multilayer combinatorial fusion for value alignment in LLMs, this paper uses F1 BERTScore as the main automatic metric for comparing generated responses with reference answers. This provides a useful first view of response quality, but it does not fully capture all aspects of value alignment. Future work should broaden the evaluation framework by incorporating additional automatic metrics such as ROUGE-L and BLEURT. More importantly, human evaluation will be essential for assessing dimensions that automatic metrics cannot fully measure. Expanding evaluation along these directions would provide a more comprehensive assessment of the strengths of MCF-CVA while helping identify where the framework can be further improved. 

\section{Conclusion}

This paper proposed the MCF-CVA framework, which integrates multi-agent moral reasoning with multilayer combinatorial fusion. By leveraging cognitive diversity among value-specific agents and applying the EAR algorithm across multiple layers, MCF-CVA dynamically refines moral judgments through both score and rank aggregation in Euclidean and Kemeny spaces.

Our results demonstrate that MCF-CVA achieves superior performance compared to single-agent and multi-agent  baselines. The multilayer fusion process enables the model to evolve toward contextually coherent and morally pluralistic decisions while maintaining diversity strength across layers. Empirical evaluations show consistent improvements in alignment quality, indicating that multilayer aggregation offers a scalable pathway toward more context-sensitive and trustworthy AI systems.

Beyond its empirical results, MCF-CVA contributes conceptually to the study of \emph{computational moral diversity}. It formalizes how moral agents with different normative perspectives can be combined into a single coherent decision-making system, providing a mathematical and algorithmic foundation for pluralistic alignment. The dual-space formulation further bridges moral reasoning and combinatorial optimization, linking ethics, information fusion, and multi-agent learning under a unified framework.

Future work will extend MCF-CVA along several directions. First, incorporating additional moral dimensions and dynamic weighting of moral foundations could enhance adaptability to novel cultural or situational contexts. Second, human-in-the-loop evaluation of moral coherence will help assess whether multilayer fusion aligns with human moral intuitions beyond static benchmarks. Finally, we aim to explore applications of MCF-CVA in collaborative agent systems and autonomous decision-making, where reconciling diverse moral reasoning is essential for collective intelligence.

Overall, MCF-CVA represents a principled step toward the operationalization of moral pluralism in large language models and offers a robust foundation for future research in multi-agent value alignment.

\bibliography{aaai2026}


\appendix

\onecolumn

\section{Contextual Value Alignment via Multilayer Combinatorial Fusion}
\vspace{0.5cm}
\section*{Appendix A: Rank-score Function and Cognitive Diversity}

\renewcommand{\thetable}{A\arabic{table}}
\setcounter{table}{0}
\renewcommand{\thefigure}{A\arabic{figure}}
\setcounter{figure}{0}

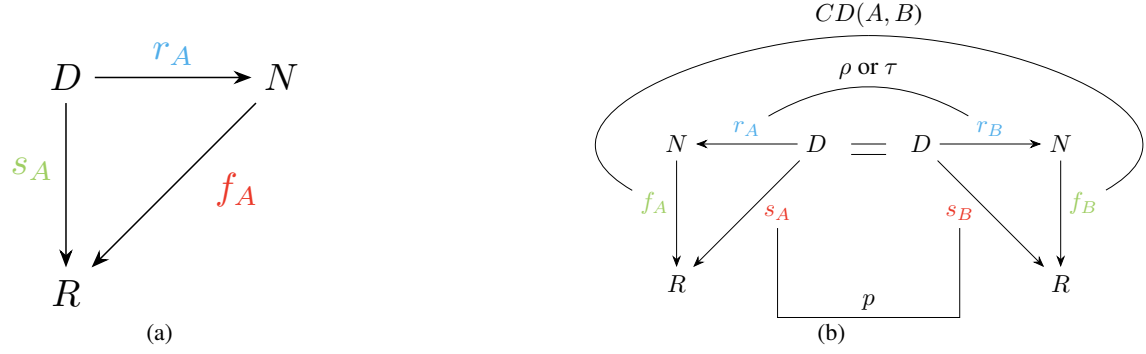
\begin{figure}[htbp]
  \centering
  \begin{subfigure}[b]{0.3\linewidth}
    \centering
    \resizebox{0.8\columnwidth}{!}{%
        \begin{tikzpicture}[every node/.style={outer sep=0pt}]
             \node[] (N) at (-2, 0) {$D$};
        \node[] (D) at (0, 0) {$N$};
        \node[] (R) at (-2, -2) {$R$};
        \draw[->, -{Stealth}] (N) -- (D) node[midway, above] {\textcolor{Blue1}{$r_A$}};
        \draw[->, -{Stealth}] (D) -- (R) node[pos=0.3, below=2pt, xshift = 8pt] {\textcolor{Red1}{$f_A$}};
        \draw[->, -{Stealth}] (N)--(R) node[pos=0.4, left] {\textcolor{Green1}{$s_A$}};
        \end{tikzpicture}
    }
    \caption{}
    \label{fig:a1}
  \end{subfigure}\hfill
  \begin{subfigure}[b]{0.7\linewidth}
    \centering
    \resizebox{0.5\columnwidth}{!}{%
        \begin{tikzpicture}[every node/.style={outer sep=0pt}, trim left = 10pt]
        \captionsetup{skip=6pt}
        \path[use as bounding box] (6,-2.5) rectangle (7,2.2);
        \node[] (D) at (5, 0) {$D$};
        \node[] (N) at (7, 0) {$N$};
        \node[] (R) at (7, -2) {$R$};
        \draw[->, -{Stealth}] (D) -- (N) node[midway, above] (rB) {\textcolor{Blue1}{$r_B$}};
        \draw[->, -{Stealth}] (D) -- (R) node[pos=0.3, below=2pt, xshift =-4pt] (sB) {\textcolor{Red1}{$s_B$}};
        \draw[->, -{Stealth}] (N)--(R) node[pos=0.4, right] (fB) {\textcolor{Green1}{$f_B$}};
        \draw[-] (4, 0) -- (4.5, 0);
        \draw[-] (4, -0.15) -- (4.5, -0.15);
        \node[] (D) at (3.5, 0) {$D$};
        \node[] (N) at (1.5, 0) {$N$};
        \node[] (R) at (1.5, -2) {$R$};
        \draw[->, -{Stealth}] (D) -- (N) node[midway, above] (rA) {\textcolor{Blue1}{$r_A$}};
        \draw[->, -{Stealth}] (D) -- (R) node[pos=0.3, below=2pt, xshift = 4pt] (sA) {\textcolor{Red1}{$s_A$}};
        \draw[->, -{Stealth}] (N)--(R) node[pos=0.4, left] (fA) {\textcolor{Green1}{$f_A$}};
    
        \draw[] (rA) to[out=30, in=150] node[midway, above] {$\rho$ or $\tau$} (rB);
        \draw[-] (sA) -- ($(sA) + (0, -1.5)$) -- ($(sB)+(0, -1.5)$) node[midway, above] {$p$} -- (sB);
        \coordinate (M) at ($(fA)!0.5!(fB) +(0, 2.4cm)$);
        \node[] at ($(M)+(0, 0.3cm)$) {$CD(A, B$)};
        \draw[] (fA) to[out=150, in=180, looseness=1.3] (M);
        \draw[] (M) to[out=0, in=30, looseness=1.3] (fB);
        \end{tikzpicture}
    }
    \caption{}
    \label{fig:b2}
  \end{subfigure}
  \caption{(a) Relationship among rank function $r_A$, score function $s_A$ and rank-score function $f_A$  \cite{hsu2006combinatorial}; (b) Cognitive diversity CD vs. Pearson's correlation ($p$), Spearman's correlation ($\rho$) and Kendall's correlation ($\tau$) \cite{hsu2019cognitive}.}
  \label{fig:two-tikz}
\end{figure}

  Under a scoring system A, score function $s_A$, its derived 
 rank function  $r_A$, and rank-score function $f_A$ as well as
their relationships are shown in Figure \ref{fig:two-tikz}(a). We note that the rank-score function $f_A$ of a scoring system A has two distinctive 
properties. It not only characterizes the scoring (or ranking) behaviour
 of the scoring system A but also provides a projection (and relation) 
between the rank function values and the score function values. \cite{hsu2002methods, hsu2006combinatorial}.
The notion of rank-score function is analogous to but different from
 the power law phenomena and rank-frequency distribution \cite{zipf2016human, cocho2019rank}.
         Figure \ref{fig:two-tikz}(b) compares cognitive diversity (CD), defined as difference between two rank-score functions $f_A$ and $f_B$, 
to three statistical data correlations: (1) Pearson correlation  ($p$)
 between score function $s_A$ and $s_B$, (2) Spearman's rho 
correlation ($\rho$) and (3) Kendall's tau correlation ($\tau$) between 
$r_A$ and $r_B$ \cite{hsu2019cognitive}. Moreover, CD is also shown 
to be analogous to but different from several diversity measures in 
computational learning and ensemble methods \cite{kuncheva2004combining, zhou2012ensemble, hsu2019cognitive, hurley2020multi}. 

\clearpage
\section*{Appendix B: Kemeny Rank Space as the Architecture for the MCF-CVA 
     Framework}

\textbf{B1: Bubble sort Cayley graph space and the permutahedron}
\vspace{0.2cm}
\label{app:groups-graphs-geometry}

\renewcommand{\thetable}{B\arabic{table}}
\setcounter{table}{0}
\renewcommand{\thefigure}{B\arabic{figure}}
\setcounter{figure}{0}

\begin{figure}[H]
    \centering
    \includegraphics[width=0.3\linewidth]{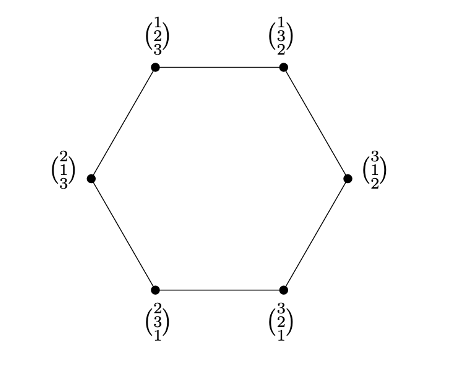}
    \caption{$B_3$ \cite{zhong2019combining}.}
    \label{fig:appA-B3}
\end{figure}

\begin{figure}[H]
    \centering
    \includegraphics[width=0.4\linewidth]{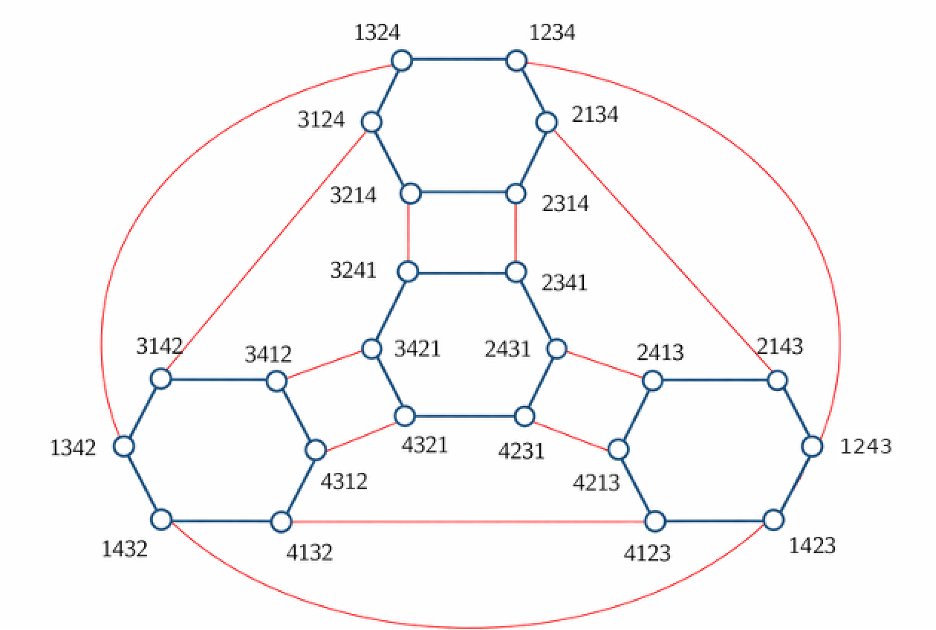}
    \includegraphics[width=0.5\linewidth]{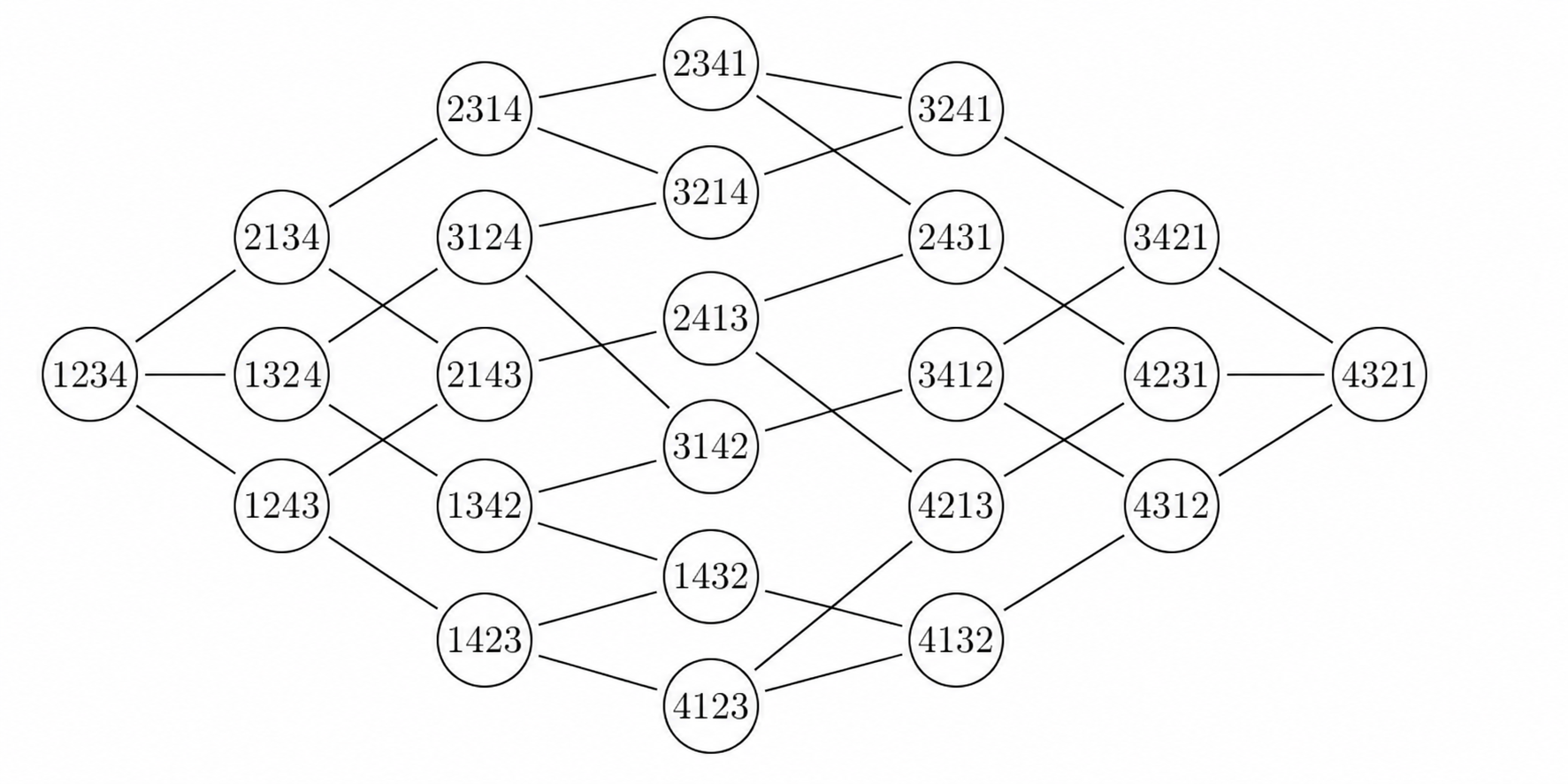}
    \caption{Left: $B_4$ (recursive); Right: $B_4$ (Cayley graph structure with generating set \{(12), (23), (34)\}) \cite{zhong2019combining}.}
    \label{fig:appA-B4}
\end{figure}

In the symmetric group $S_n$, let $T_n$ be the subset of all $n-1$ adjacent transpositions $\{(1\,2),(2\,3),\ldots,(n-1\,n)\}$. The Cayley graph $\mathrm{Cay}(S_n,T_n)$ is constructed as the graph with $S_n$ as the vertex set and
\(\{(\pi,\pi\circ t)\}\) as the edge set, where $\pi\in S_n$ and $t\in T_n$ \cite{diaconis1988group}.
This graph $B_n=\mathrm{Cay}(S_n,T_n)$ is also called Bubble Sort Cayley Graph because any pair of vertices
$A$ and $B$ are connected by a path of distance equal to the number of adjacent swaps by the sorting
algorithm: bubble sort. The graph $B_n$ has many combinatorial and graphical properties. For example, it is $(n-1)$-regular because the generating set $T_n$ has $n-1$ elements; it is bipartite and has diameter $n(n-1)/2$.
Moreover, graph $B_{n+1}$ can be recursively constructed by $n+1$ copies of $B_n$.
Figure~\ref{fig:appA-B3} illustrates the graph $B_3$ and Figure \ref{fig:appA-B4} shows the two representations of the graph of $B_4$.

\begin{figure}[H]

    \centering
    \includegraphics[width=0.35\linewidth]{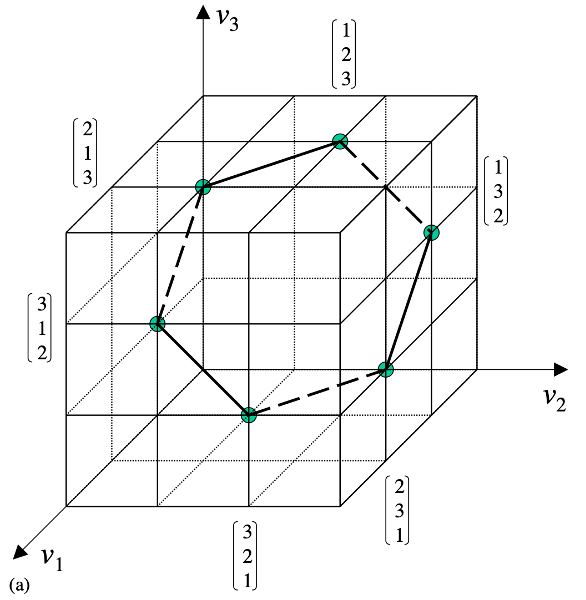}
    \hspace{1cm}
    \includegraphics[width=0.28\linewidth]{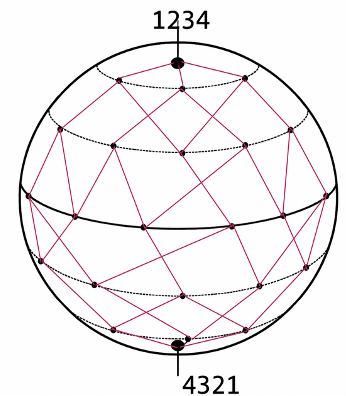}
    \caption{Left: $P_3$ \cite{zhang2004binary}; Right: $P_4$ \cite{marmor2021accuracy}.}
    \label{fig:appA-permutahedra}
\end{figure}

If we treat each permutation (or rankings without ties) as a vector, the set of all permutations on the $n$ elements
constitutes the convex hull of a permutahedron $P_n$ \cite{zhang2004binary}. Considering each permutation of the set of data items $D$, $\lvert D \rvert = n$,  as a preference chain of these $n$ items, the geodesic distance metric (GDM) of dissimilarity (or diversity) between two
preference chains is proportional to the angle and hence the length of the geodesic arc connecting the two
corresponding points on the surface of the $n$-dimensional sphere. \citet{marmor2021accuracy} showed that this GDM
satisfies the requirements of a distance metric, including the triangle inequality, and is related to Kendall's
$\tau$ correlation coefficient of similarity. Examples of $P_3$ and $P_4$ are illustrated in
Figure~\ref{fig:appA-permutahedra}.

\noindent\textbf{B2: Kemeny rank space}
\vspace{0.2cm}

When the score function $s_A$ of a scoring system $A$ is not 1-1 function, the derived rank function is not a
permutation. This is the case when ``tie ranking'' occurs. Although there are several methods to resolve tie
ranking situation such as taking the average rank, Kendall's $\tau$ correlation distance does not apply to tie rankings. \citet{kemeny1962mathematical} proposed a distance metric $d_K$ that can handle ties. The $d_K$ metric was updated by \citet{emond2002new} to a new rank correlation $\tau_x$ between two weak orders (rankings with
ties) as inner product of their score function values due to efficient computation.

\begin{figure}[H]
    \centering
    \includegraphics[width=0.35\linewidth]{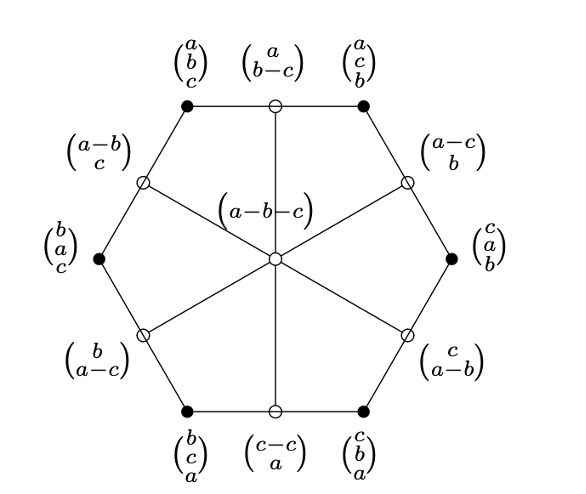}
    \caption{Kemeny rank space $K_3$ \cite{kemeny1962preference}.}
    \label{fig:appA-K3}
\end{figure}

Since the Kemeny rank space $K_n$ consists of all rankings (ties or no ties) of the dataset items
$\{d_1,d_2,\ldots,d_n\}$, the vertex set of the Kemeny rank space $K_n$ has more vertices than the bubble-sort Cayley graph space $B_n$, which is $n!$. The number of vertices $f(n)$ of the Kemeny rank space $K_n$ has been calculated from two perspectives. \citet{good1975number} showed that $f(n)=\sum_{b=1}^{n} b!\, S(n,b)$, where $S(n,b)$ is the Stirling number of the second kind. On the other hand, $f(n)$ was obtained by \citet{gross1962preferential} as a recursive formula
$f(n)=1+\sum_{i=1}^{n} \binom{n}{i}\, f(n-i)$ in terms of preference arrangements w.r.t. $n$ objects allowing indifferences. The Kemeny rank space $K_3$ is
illustrated in Figure~\ref{fig:appA-K3}.

The Kemeny rank space $K_n$ has been used in information retrieval, multi-criteria decision making, and computational
social science \cite{akbari2023beyond}. More recently, focus has been on rank aggregation on the Kemeny space, in which a derived median ranking is pursued in terms of optimizing computational complexity \cite{akbari2023beyond}. In our application, we perform multi-agent multilayer combinatorial fusion (MCF) on the Kemeny rank space using both average and weighted combinations of multi-agents' rankings.

\newpage
\section*{Appendix C: Figure Skating Judgement}
\vspace{0.5cm}

\renewcommand{\thetable}{C\arabic{table}}
\setcounter{table}{0}
\renewcommand{\thefigure}{C\arabic{figure}}
\setcounter{figure}{0}

In this section, we use an example in figure skating judgement 
to illustrate (a) score combination v.s. rank combination
 and (b) rank-score function and comparison of rank-score function 
graphs w.r.t. judges $J_1, J_2$, and $J_3$ \cite{jiang2023enhancing}. In the judgement of figure skating competition, suppose we would like to rank a group of 8 skaters by 3 judges. The set of skaters $D = \{ d_1, d_2, \ldots, d_8 \}$,
and the set of judges is $\{ J_1, J_2, J_3 \}.$ The following tables C1, C2, C3 and C4 illustrate the score functions $s_{J_1}$, $s_{J_2}$, and $s_{J_3}$ and their score combinations; the rank functions $r_{J_1}$, $r_{J_2}$, and $r_{J_3}$ and their rank combinations; the normalized score values; and the rank--score functions $f_{J_1}$, $f_{J_2}$, and $f_{J_3}$, respectively. Figure C1 illustrates the rank--score function graphs. The cognitive diversity between scoring systems A and B is the area between these two graphs.

Note that this example demonstrates that judge $J_3$ manipulates the score on skater $d_8$. Table C2 indicates how to mitigate the problem while Figure C1 shows how to detect the bias by exhibiting rank--score function graphs of judges~$J_1$, $J_2$, and~$J_3$.

\begin{figure}[H]
\centering

\begin{minipage}[t]{0.49\linewidth}
\centering
\captionof{table}{Score function and score combination.}
\setlength{\tabcolsep}{7pt}
\begin{tabular}{cccccc}
\toprule
      & $s_{J1}$  & $s_{J2}$  & $s_{J3}$  & $s_{SC}$   & $r_{SC}$ \\
\midrule
$d_1$ & 9.6 & 9.7 & 9.8 & 29.1  &2\\
$d_2$ & 9.8 & 9.2 & 9.9 & 28.9 &3 \\
$d_3$ & 9.7 & 9.9 & 10  & 29.6 &1\\
$d_4$ & 9.5 & 9.3 & 9.7 & 28.5 &6\\
$d_5$ & 9.9 & 9.4 & 9.5 & 28.8 &4\\
$d_6$ & 9.4 & 9.6 & 9.6 & 28.6 &5\\
$d_7$ & 9.3 & 9.5 & 9.4 & 28.2 &7\\
$d_8$ & 10  & 10  & \textbf{7}   & 27.0 & \textbf{8}\\
\bottomrule
\end{tabular}
\end{minipage}
\hfill
\begin{minipage}[t]{0.49\linewidth}
\centering
\captionof{table}{Rank function and rank combination.}
\setlength{\tabcolsep}{7pt}
\begin{tabular}{cccccc}
\toprule
      & $r_{J1}$ & $r_{J2}$ & $r_{J3}$ & $s_{RC}$ & $r_{RC}$ \\
\midrule
$d_1$ & 5  & 3  & 3  & 11 & 3 \\
$d_2$ & 3  & 8  & 2  & 13 & 4 \\
$d_3$ & 4  & 2  & 1  & 7  & 1 \\
$d_4$ & 6  & 7  & 4  & 17 & 7 \\
$d_5$ & 2  & 6  & 6  & 14 & 5 \\
$d_6$ & 7  & 4  & 5  & 16 & 6 \\
$d_7$ & 8  & 5  & 7  & 20 & 8 \\
$d_8$ & 1  & 1  & 8  & 10 & \textbf{2} \\
\bottomrule
\end{tabular}
\end{minipage}

\vspace{1.2em}

\begin{minipage}[t]{0.49\linewidth}
\centering
\captionof{table}{Normalized score values.}
\setlength{\tabcolsep}{7pt}
\begin{tabular}{cccc}
\toprule
      & $s_{J1}$   & $s_{J2}$   & $s_{J3}$   \\
\midrule
$d_1$ & 0.43 & 0.63 & 0.93 \\
$d_2$ & 0.71 & 0.00 & 0.97 \\
$d_3$ & 0.57 & 0.75 & 1.00 \\
$d_4$ & 0.28 & 0.13 & 0.90 \\
$d_5$ & 0.86 & 0.25 & 0.83 \\
$d_6$ & 0.14 & 0.50 & 0.86 \\
$d_7$ & 0.00 & 0.38 & 0.80 \\
$d_8$ & 1.00 & 1.00 & 0.00 \\
\bottomrule
\end{tabular}
\end{minipage}
\hfill
\begin{minipage}[t]{0.49\linewidth}
\centering
\captionof{table}{Rank-score functions $f_{J_1}, f_{J_2}, f_{J_3}$.}
\setlength{\tabcolsep}{7pt}
\begin{tabular}{cccc}
\toprule
Rank & $f_{J_1}$ & $f_{J_2}$ & $f_{J_3}$ \\
\midrule
1 & 1.00 & 1.00 & 1.00 \\
2 & 0.86 & 0.75 & 0.97 \\
3 & 0.71 & 0.63 & 0.93 \\
4 & 0.57 & 0.50 & 0.90 \\
5 & 0.43 & 0.38 & 0.86 \\
6 & 0.28 & 0.25 & 0.83 \\
7 & 0.14 & 0.13 & 0.80 \\
8 & 0.00 & 0.00 & 0.00 \\
\bottomrule
\end{tabular}
\end{minipage}

\vspace{1.2em}

\begin{minipage}[t]{\linewidth}
\centering
\captionof{figure}{Rank--score function graphs.}
\begin{tikzpicture}
\begin{axis}[
    width=0.6\linewidth,
    height=6cm,
    xmin=1, xmax=8,
    ymin=0, ymax=1.05,
    xlabel={Rank},
    ylabel={Normalized Score},
    xtick={1,2,3,4,5,6,7,8},
    ytick={0,0.1,...,1.0},
    grid=major,
    legend style={at={(1.02,0.5)}, anchor=west},
]
\addplot+[mark=square*, line width = 2pt] coordinates {
    (1,1.00) (2,0.86) (3,0.71) (4,0.57) (5,0.43) (6,0.28) (7,0.14) (8,0.00)
};
\addlegendentry{$f_{J_1}$}

\addplot+[mark=square*,   line width = 2pt] coordinates {
    (1,1.00) (2,0.75) (3,0.63) (4,0.50) (5,0.38) (6,0.25) (7,0.13) (8,0.00)
};
\addlegendentry{$f_{J_2}$}

\addplot+[mark=square*,   line width = 2pt] coordinates {
    (1,1.00) (2,0.97) (3,0.93) (4,0.90) (5,0.86) (6,0.83) (7,0.80) (8,0.00)
};
\addlegendentry{$f_{J_3}$}
\end{axis}
\end{tikzpicture}
\end{minipage}

\end{figure}

\newpage
\section*{Appendix D: EAR Algorithm and the MCF-CVA Framework}
\vspace{0.5cm}

\renewcommand{\thealgorithm}{D\arabic{algorithm}}
\setcounter{algorithm}{0}

\begin{algorithm}[hp]
\caption{Expansion-and-Reduction (EAR)}\label{alg:er}
\label{alg:muf}
\begin{algorithmic}[1]
\Require $t$ scoring systems $\mathcal{M}=\{M_1,\dots,M_t\}$ and set of data items $D = \{d_1, d_2, \cdots, d_n\}$.
\Ensure New $t$ scoring systems $\mathcal{M}^\prime=\{M_1^\prime,\dots,M_t^\prime\}$ for next layer.

    \For{$k = 2, 3, \cdots, t$} \Comment{$k$ models to combine: 2-com, 3-com, $\cdots$, $t-$com.}
      \ForAll{$I \in \Call{Choose}{\{1,\ldots,t\},\,k}$}. \Comment{Select which $k$ scoring systems for combination.}

      \State $\{\mathcal{S}_i : i \in I\} \gets \text{Get score function for each system }{\{\mathcal{M}_i : i \in I\}}$.
      \State $S_I^{\text{\tiny ASC}} \gets \text{Do average score combination for }{\{\mathcal{S}_i : i \in I\}}$ using formula (\ref{eq:formula 1}).
      \State \text{Compute cognitive diversity for every pair from }${\{\mathcal{S}_i : i \in I\}}$.
      \State $DS \gets \text{Compute diversity strength for each system}$.
      \State $S_I^{\text{\tiny WSCDS}} \gets \text{Do weighted score combination by DS using formula}$ (\ref{eq:formula 2}).
    
      \State $\{\mathcal{R}_i : i \in I\} \gets \text{Get rank function for each system}{\{\mathcal{M}_i : i \in I\}}$. 
      \State $S_I^{\text{\tiny ARC}} \gets \text{Do average rank combination for }{\{\mathcal{R}_i : i \in I\}}$ using formula (\ref{eq:formula 3}).
      \State $S_I^{\text{\tiny WRCDS}}   \gets \text{Do weighted rank combination by DS}$ using formula (\ref{eq:formula 4}).
      \EndFor
    \EndFor

    \State  $\mathcal{S}_{n\times \big(4(2^t-t-1)\big)} \gets \text{Horizontally stack} {\{\, S_I^{\mathrm{ASC}}, S_I^{\mathrm{WSCDS}},
    S_I^{\mathrm{SARC}}, S_I^{\mathrm{WRCDS}}\}}$ \text{together}. \Comment{Each fusion type has $2^t-t-1$ combs.}
    
    \State $DS_{1\times \big(4(2^t-t-1)\big)} \gets \text{Compute diversity strength for each column of }{\mathcal{S}_{n\times \big(4(2^t-t-1)\big)}}$. 
    
     \State $ \{\mathcal M_1^\prime,\dots,\mathcal M_t\prime\} \gets \text{Obtain top } t \text{ systems from }{ \mathcal{S}_{n\times \big(4(2^t-t-1)\big)}}$ based on $DS_{1\times \big(4(2^t-t-1)\big)}$. \Comment{Sort all systems by DS in descending order and take the top $t$ systems.}

\State \Return $\{\mathcal M_1^\prime,\dots,\mathcal M_t^\prime\}$
\end{algorithmic}
\end{algorithm}

\newpage

\vspace{2cm}
\begin{algorithm}[H]
\caption{Multilayer combinatorial fusion for contextual value alignment (MCF-CVA)}\label{alg:mcf4va}
\begin{algorithmic}
\Require Five fine-tuned moral agents $\mathcal{A}=\{A_1,\dots,A_5\}$; test question set $\mathcal{Q}$.
\Ensure The F1 BERTScore at final layer, $\mathtt{MCFperf}$ and the last layer \(\ell^{\ast}\) at which the stopping criterion is met.

\ForAll{$q \in \mathcal{Q}$}                                   \Comment{Per question.}
  \For{each moral agent $A_i \in \mathcal{A}$}                                    
    \State $r_i \gets \text{Generate a response for }{\text{agent }A_i \text{ on question } q}$.
    \State $U_i \gets \text{Partition } r_i \text{ into moral units} \text{ using } \texttt{GPT-4.1-nano}$.
  \EndFor
  \State $U \gets$  Pool all moral units $U_1 \cup U_2 \cup \cdots \cup U_5$.    
  \State $\langle d_1,\dots,d_n\rangle \gets$ Form all possible unit pairs $\{(u_i,u_j)\mid u_i,u_j\in U,\ i<j\}$ as data items.

    \State $\text{ST} \gets \texttt{SentenceTransformer}(\texttt{all-MiniLM-L6-v2})$.
    \For{$k \gets$  Each data item (unit pair) $1$ to $n$}
        \State $\text{Encode each data item to tokens for }{d_k}$ using ST. \Comment{Features for each pair.}
        \State $\hat{p}_k \gets \text{Train Logistic Regression and get predictions for } d_k$.  \Comment{Pre-trained multi-label classifier.}
    \EndFor

\State $\{\mathcal S_1,\dots,\mathcal S_5\} \gets
       \text{Build five scoring systems}{\{\hat p_k\}_{k=1}^{n}}$. \Comment{Each aligned with one moral value.}

\State Initialize $\ell = 0$.  \Comment{Layer 0.}
\While{$\Call{StoppingCriteria}{\{\mathcal M_1,\dots,\mathcal M_5\}}$} \Comment{Maximum diversity strength $\geq 0.05$ and number of layer $\leq$ 6.}
    \State $\ell \gets \ell+1$.      \Comment{Layer increases.}                             
    \State $\{\mathcal M_1,\dots,\mathcal M_5\} \gets \Call{Expansion-and-Reduction}{\{\mathcal M_1,\dots,\mathcal M_5\}}$. \Comment{From Algorithm \ref{alg:er}.}
\EndWhile
\State $\ell^\ast(q) \gets \ell$. \Comment{Store the number of layer for question $q$.}
\State Average rank combination of $\{\mathcal M_1,\dots,\mathcal M_5\}$. \Comment{Get final ranking of data items (unit pair).}
\State Obtain best data item(s). \Comment{Sort ranks ascendingly to get the top data item(s).}
\State $\mathtt{MCFperf}(q) \gets $ Aggregating the data item(s).

\EndFor 
\State \Return $\langle \mathtt{MCFperf},\   \ell^\ast\rangle$
\end{algorithmic}
\end{algorithm}
\clearpage

\section*{Appendix E: Perfomance Summary w.r.t. Question $\boldsymbol{q_{2835}}$}
\vspace{0.5cm}
\renewcommand{\thetable}{E\arabic{table}}
\setcounter{table}{0}
\renewcommand{\thefigure}{E\arabic{figure}}
\setcounter{figure}{0}

\begin{table}[H]
    \centering
    \caption{Performance summary for the best single models and the best combined models for layer 1,2,3,4, 5 and 6 on question $q_{2835}$, respectively (\texttt{a}: ASC; \texttt{b}: WSCDS; \texttt{c}: ASC\&WSCDS; \texttt{d}: ARC, \texttt{e}: WRCDS; \texttt{f}: ARC\&WRCDS; \texttt{g}: all four types of combination.)}
    \begin{tabular}{c|l|p{7cm}|c}
    \hline
       \#Layer  &  Model type &   Best model  & F1 BERTScore \\
       \hline
      \multirow{2}{*}{Layer 1} & single & C & \textbf{0.8663}\\
      & combined & AC (\texttt{d}), ACE (\texttt{d}), ABCDE (\texttt{d}) & 0.8756	\\ 
      \hline
      \multirow{2}{*}{Layer 2}  & single & C, E & 0.8628\\
      & combined & AC (\texttt{f}), CD (\texttt{g}), ACD (\texttt{d})  & 0.8698\\ 
      \hline
    \multirow{2}{*}{Layer 3} & single  & A & 0.8698 \\ 
     & combined & ABC (\texttt{d}), ACE (\texttt{f}) & 0.8708\\ 
      \hline
    \multirow{2}{*}{Layer 4} & single  & A & 0.8698 \\ 
     & combined & AB (\texttt{f}), ABD (\texttt{g}), BDE (\texttt{a}) & 0.8755\\ 
      \hline
        \multirow{2}{*}{Layer 5} & single  & B & 0.8698 \\ 
     & combined & CE (\texttt{f}), DE (\texttt{d}), ADE (\texttt{a}) & 0.8755\\ 
      \hline
        \multirow{2}{*}{Layer 6} & single  & A & 0.8698 \\ 
     & combined & AE (\texttt{g}), ACE (\texttt{f}), ADE (\texttt{d}) & \textbf{0.8797}\\ 
      \hline
    \end{tabular}
    
    \label{tab:bestperf7049}
\end{table}

\begin{figure}[H]
    \centering
    \includegraphics[width=0.3\linewidth]{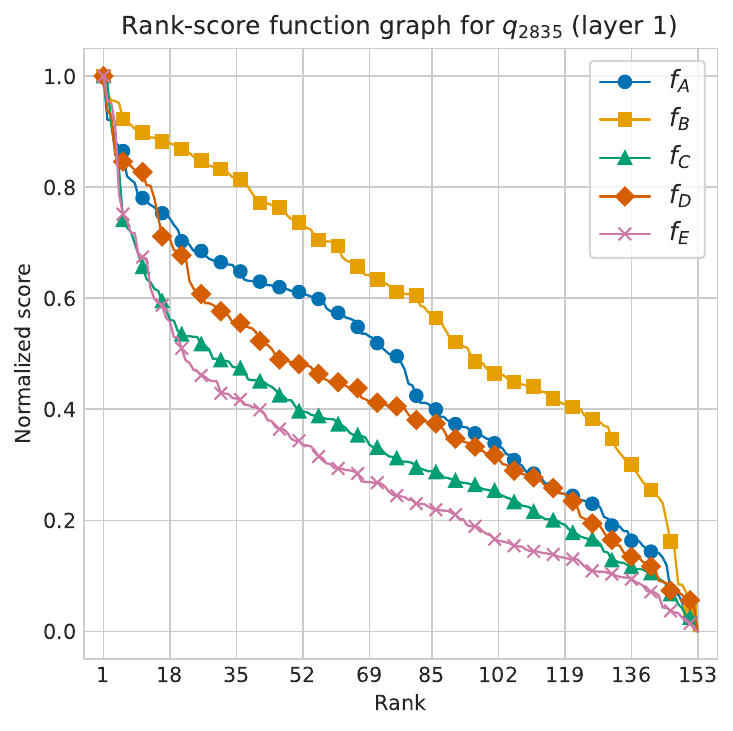}
    \includegraphics[width=0.3\linewidth]{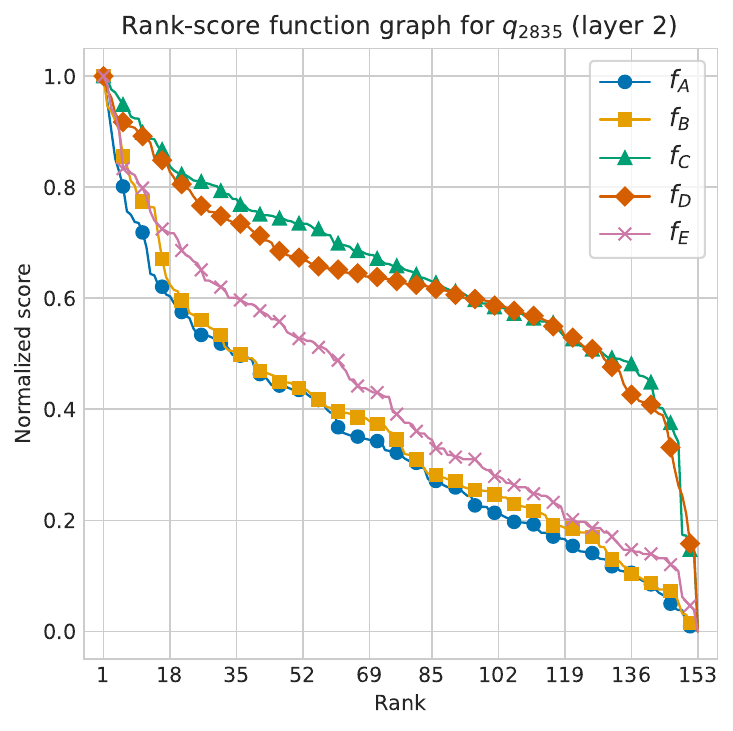}
    \includegraphics[width=0.3\linewidth]{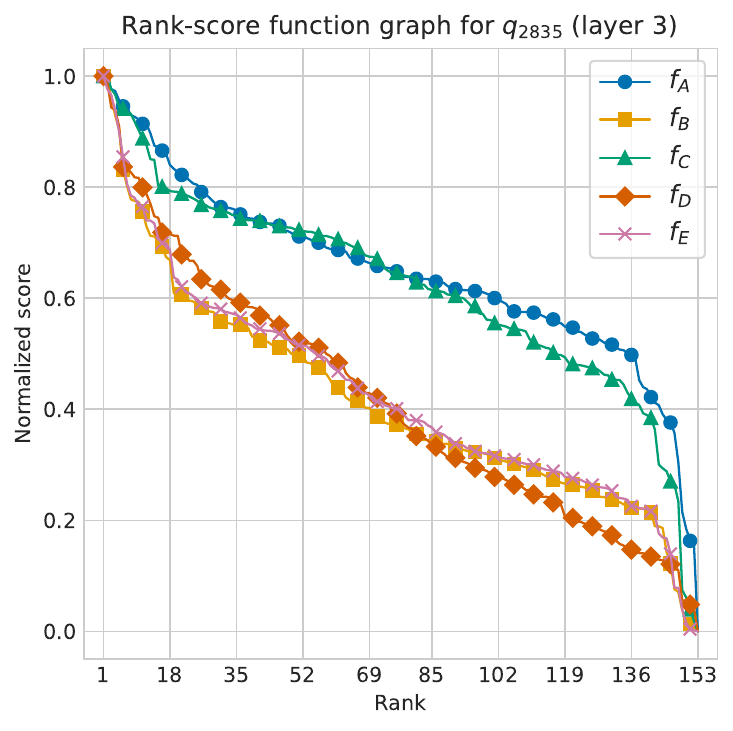}
    \includegraphics[width=0.3\linewidth]{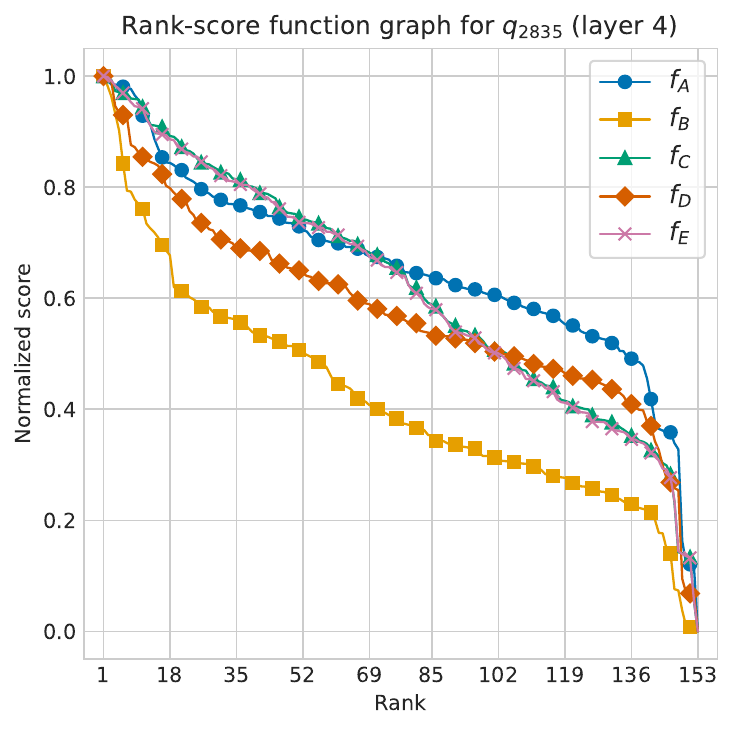}
    \includegraphics[width=0.3\linewidth]{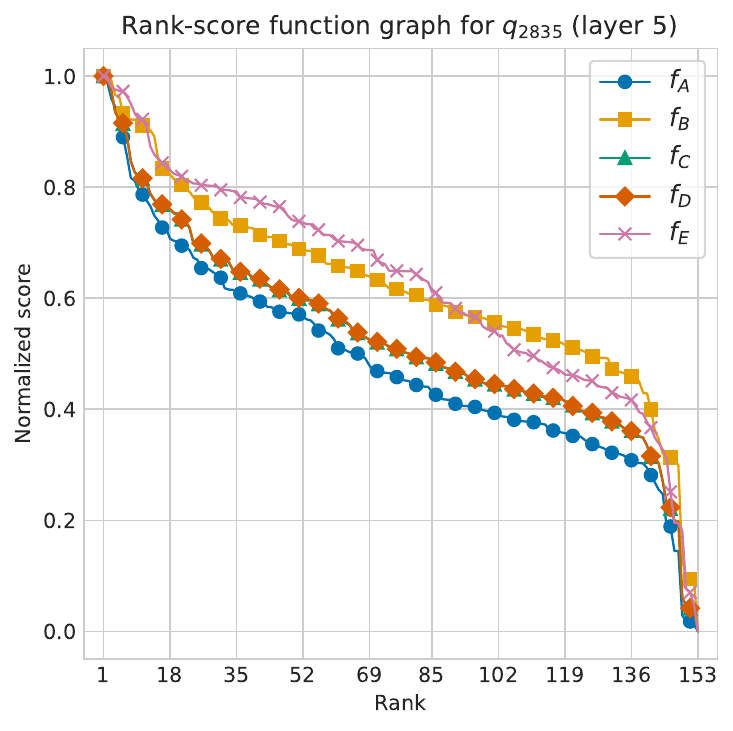}
    \includegraphics[width=0.3\linewidth]{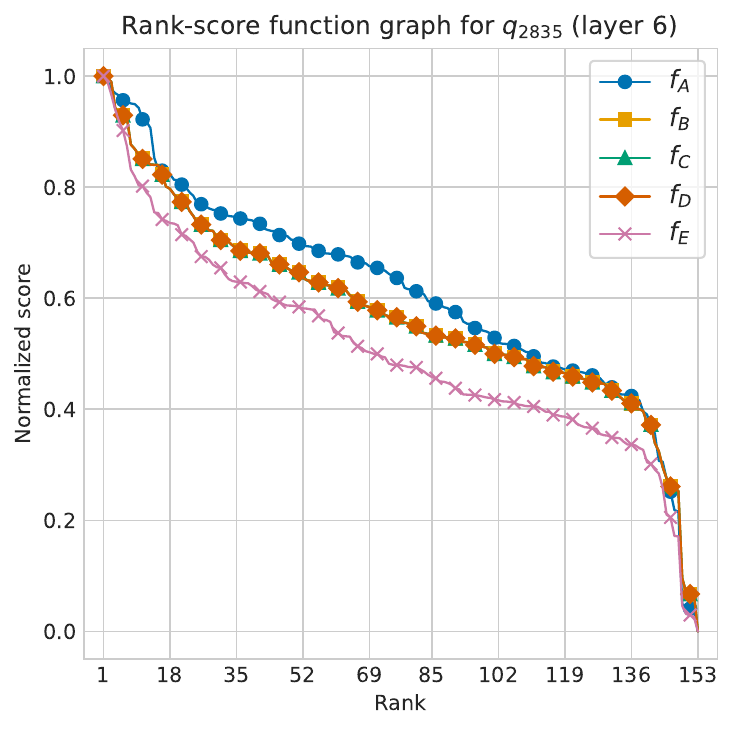}
    \caption{Rank-score function graph for 6 layers w.r.t question $q_{2835}$.}
    \label{fig:placeholder}
\end{figure}

\newpage

\begin{landscape}
\section*{Appendix F: A Detailed Example of Response Decomposition in MCF-CVA Workflow}
\vspace{0.5cm}
\renewcommand{\thetable}{F\arabic{table}}
\setcounter{table}{0}
\renewcommand{\thefigure}{F\arabic{figure}}
\setcounter{figure}{0}

\begin{figure}[H]
\centering

\begin{tikzpicture}[
  font=\small,
  box/.style={draw, rounded corners=3pt, align=center, minimum height=1.2cm},
  bigbox/.style={box, minimum width=23cm, text width = 22cm, align = left},
  smallbox/.style={box, minimum width=4.2cm, minimum height=5.5cm,   align=left, text width=4cm},
  gap/.style={},
]

\node[bigbox] (top) {\textbf{Question:} If we never started to globally recycle in the 90s, how bad would it be right now and would the world look any different?};

\node[bigbox, below=0.2cm of top] (responses) {\textbf{Moral agent responses} \\
\textbf{Moral agent A (Authority):} If global recycling had never begun in the 1990s, governments and institutions would have faced greater difficulty maintaining environmental order and public standards. The lack of coordinated recycling policies would weaken regulatory authority, leading to more uncontrolled waste and environmental degradation.\\
\textbf{Moral agent B (Care):} Recycling has been beneficial to people and other living beings, reducing pollution and helping protect the places they depend on. Without recycling, there would likely be much more pollution today, causing greater harm to communities, wildlife, and ecosystems. \\
\textbf{Moral agent C (Fairness):} If we never started to globally recycle in the 90s, the world
would be worse off in how environmental costs and benefits are shared. The amount of waste and pollution would
be much higher, and there would be fewer efforts to conserve our shared environment.\\
\textbf{Moral agent D (Loyalty):} If global recycling had never started in the 1990s, societies would show less collective responsibility toward protecting shared environments. Communities would be less united in caring for common resources, weakening cooperation across nations and groups.\\
\textbf{Moral agent E (Sanctity):} If global recycling had never begun in the 1990s, the natural world would be treated with less respect and care. Pollution and waste would accumulate more heavily, degrading the purity of land and water. This neglect would reflect a failure to protect the integrity of the environment.
};

\node[smallbox, below=0.3cm of responses, xshift=-9.2cm] (b1) {
\textbf{A1.} Without global recycling, governments and institutions would have weaker control over environmental regulation and waste management. \\
\textbf{A2.} The absence of coordinated recycling policies would undermine public order and environmental standards enforced by authorities.};
\node[smallbox, right=0.35cm of b1] (b2) {\textbf{B1.} Recycling has been beneficial to people and other living beings. \\ \textbf{B2.} Recycling reduces pollution and helps protect the
environments. \\ \textbf{B3.} If recycling had not been widely
adopted, pollution levels today would likely be much higher. \\ \textbf{B4.} Higher pollution levels
would cause greater harm to human communities, wildlife, and ecosystems. };
\node[smallbox, right=0.35cm of b2] (b3) {\textbf{C1.} The absence of global
recycling since the 1990s would make the distribution of environmental costs and benefits more unfair. \\ \textbf{C2.}
Without global recycling, the amount of waste and pollution would be much higher. \\ \textbf{C3.}
Global recycling programs help conserve the shared environment by reducing waste and reusing materials.};
\node[smallbox, right=0.35cm of b3] (b4) {
\textbf{D1.} Global recycling encourages collective responsibility toward protecting shared environmental resources.\\
\textbf{D2.} Without recycling, cooperation and solidarity among communities and nations in environmental protection would be weaker.};
\node[smallbox, right=0.35cm of b4] (b5) {
\textbf{E1.} Without recycling, pollution and waste would accumulate, degrading the purity of land and water.\\
\textbf{E2.} Recycling reflects respect for the natural world and its inherent cleanliness and integrity.\\
\textbf{E3.} Neglecting recycling represents a failure to protect the sanctity of the environment from contamination.};

\node[bigbox, below=0.2cm of b3] (mid) {\textbf{Unit pair pool:}  $\binom{14}{2} = 91$ unit pairs for the base units \{ A1, A2, B1, B2, B3, B4, C1, C2, C3, D1, D2, E1, E2, E3 \}.\\
\textbf{Best unit pair} \textbf{(B3, C3):} If recycling had not been widely adopted, pollution levels today would likely be much higher. Global recycling programs help conserve the shared environment by reducing waste and reusing materials. \\
\textbf{Aggregation:} If we had never started global recycling in the 90s,
pollution levels today would likely be much higher, because recycling programs help conserve our shared
environment by reducing waste and reusing materials. \\
\textbf{Ground truth: }There could be a lot of pollution. Recycling helps conserve the environment.
};

\end{tikzpicture}
\caption{A detailed example of response decomposition in MCF-CVA workflow.}
\end{figure}

\end{landscape}

\end{document}